\documentclass[journal]{IEEEtran}
\usepackage[utf8]{inputenc}
\usepackage{todonotes}
\usepackage{siunitx}
\usepackage[colorlinks=true, allcolors=black]{hyperref}
\usepackage{amsmath,amssymb}
\usepackage{amsfonts}
\usepackage{soul} 
\usepackage{graphicx}
\usepackage{subcaption}
\usepackage{bbding}
\usepackage{pifont}
\usepackage{algorithmicx,algorithm}
\usepackage{multirow}
\usepackage{booktabs} 
\usepackage{makecell}
\usepackage{stfloats} 
\usepackage{sidecap}

\usepackage{tikz,xcolor}
\usepackage{pgfplots}
\usepackage{filecontents}
\usepackage{color, colortbl}
\usepackage{diagbox}
\usepackage{caption}
\usepackage{enumitem}

\definecolor{lime}{HTML}{A6CE39}
\DeclareRobustCommand{\orcidicon}
{
    \begin{tikzpicture}
    \draw[lime, fill=lime] (0,0) circle [radius=0.16] 
    node[white] {{\fontfamily{qag}\selectfont \tiny ID}};    \draw[white, fill=white] (-0.0625,0.095) circle [radius=0.007];    
    \end{tikzpicture}
    \hspace{0mm}}
\foreach \x in {A, ..., Z}{%
    \expandafter\xdef\csname orcid\x\endcsname{\noexpand\href{https://orcid.org/\csname orcidauthor\x\endcsname}{\noexpand\orcidicon}}
}

\begin{document}

\title{SUSTechGAN: Image Generation for Object Detection in Adverse Conditions of \\ Autonomous Driving}

\author{
\vspace{-10pt} 
Gongjin Lan\hspace{-1mm}\orcidA{}\hspace{-1mm}, \IEEEmembership{Member, IEEE}, Yang Peng, Qi Hao \hspace{-1mm}\Envelope\hspace{-1mm}\orcidB{}\hspace{-1mm}, \IEEEmembership{Member, IEEE}, Chengzhong Xu, \IEEEmembership{Fellow, IEEE} 
\thanks{
This work was supported in part by the National Natural Science Foundation of China (62261160654), in part by the Shenzhen Fundamental Research Program (JCYJ20220818103006012,KJZD20231023092600001), in part by the Shenzhen Key Laboratory of Robotics and Computer Vision (ZDSYS20220330160557001), in part by the GuangDong Basic and Applied Basic Research Foundation (2021A1515110641), in part by the Science and Technology Development Fund of Macao S.A.R (FDCT) (No. 0123/2022/AFJ and No. 0081/2022/A2). (Corresponding author: Qi Hao.)
}
\thanks{Gongjin Lan and Yang Peng are with the Department of Computer Science and Engineering, Southern University of Science and Technology, China}
\thanks{Qi Hao is with the Research Institute of Trustworthy Autonomous Systems, and the Department of Computer Science and Engineering, Southern University of Science and Technology, China. (e-mail: haoq@sustech.edu.cn)}
\thanks{Chengzhong Xu is with the University of Macau, Macao SAR, China}
\vspace*{-28pt}
}

\markboth{Journal of \LaTeX\ Class Files,~Vol.
}%
{Shell \MakeLowercase{\textit{et al.}}: Bare Demo of IEEEtran.cls for IEEE Journals}

\maketitle

\begin{abstract}
Autonomous driving significantly benefits from data-driven deep neural networks.
However, the data in autonomous driving typically fits the long-tailed distribution, in which the critical driving data in adverse conditions is hard to collect.
Although generative adversarial networks (GANs) have been applied to augment data for autonomous driving, generating driving images in adverse conditions is still challenging. 
In this work, we propose a novel framework, SUSTechGAN, with customized dual attention modules, multi-scale generators, and a novel loss function to generate driving images for improving object detection of autonomous driving in adverse conditions.
We test the SUSTechGAN and the well-known GANs to generate driving images in adverse conditions of rain and night and apply the generated images to retrain object detection networks. 
Specifically, we add generated images into the training datasets to retrain the well-known YOLOv5 and evaluate the improvement of the retrained YOLOv5 for object detection in adverse conditions.
The experimental results show that the generated driving images by our SUSTechGAN significantly improved the performance of retrained YOLOv5 in rain and night conditions, which outperforms the well-known GANs.  
The open-source code, video description and datasets are available on the page \footnote{\url{https://github.com/sustech-isus/SUSTechGAN}} to facilitate image generation development in autonomous driving under adverse conditions.
\end{abstract}


%
\IEEEpeerreviewmaketitle

\vspace{-8pt}
\section{Introduction}
\vspace{-3pt}


Autonomous driving generally refers to an SAE J3016 
\footnote{\url{https://www.sae.org/blog/sae-j3016-update}} level 3 or higher technology system that controls the vehicle to reach the destination by detecting the external environment without driver intervention.
However, fully autonomous driving would have to be driven hundreds of millions of miles and even hundreds of billions of miles to collect data and demonstrate their safety \cite{kalra2016driving}.
In the real world, the existing fleets would take tens and even hundreds of years to drive these miles which is significantly time-consuming and expensive, even impossible.
Furthermore, data labeling is a significantly challenging and heavy workload.
Finally, collecting real driving data under adverse conditions is prominently challenging and costly.
Although data collection in simulation could collect large-scale data to provide ground truth and improve autonomous driving, 
the reality gap between simulated data and the real data is a significant issue.
Data augmentation by generative AI has been demonstrated to reduce road testing costs and improve autonomous vehicle safety.
GAN-based methods have been successfully applied to generate driving data under adverse conditions which are significantly difficult to collect in the real world \cite{huang2018auggan}.

\begin{figure*} [!ht] \centering
  \includegraphics[width=0.93\linewidth,trim={170 241 180 70},clip]{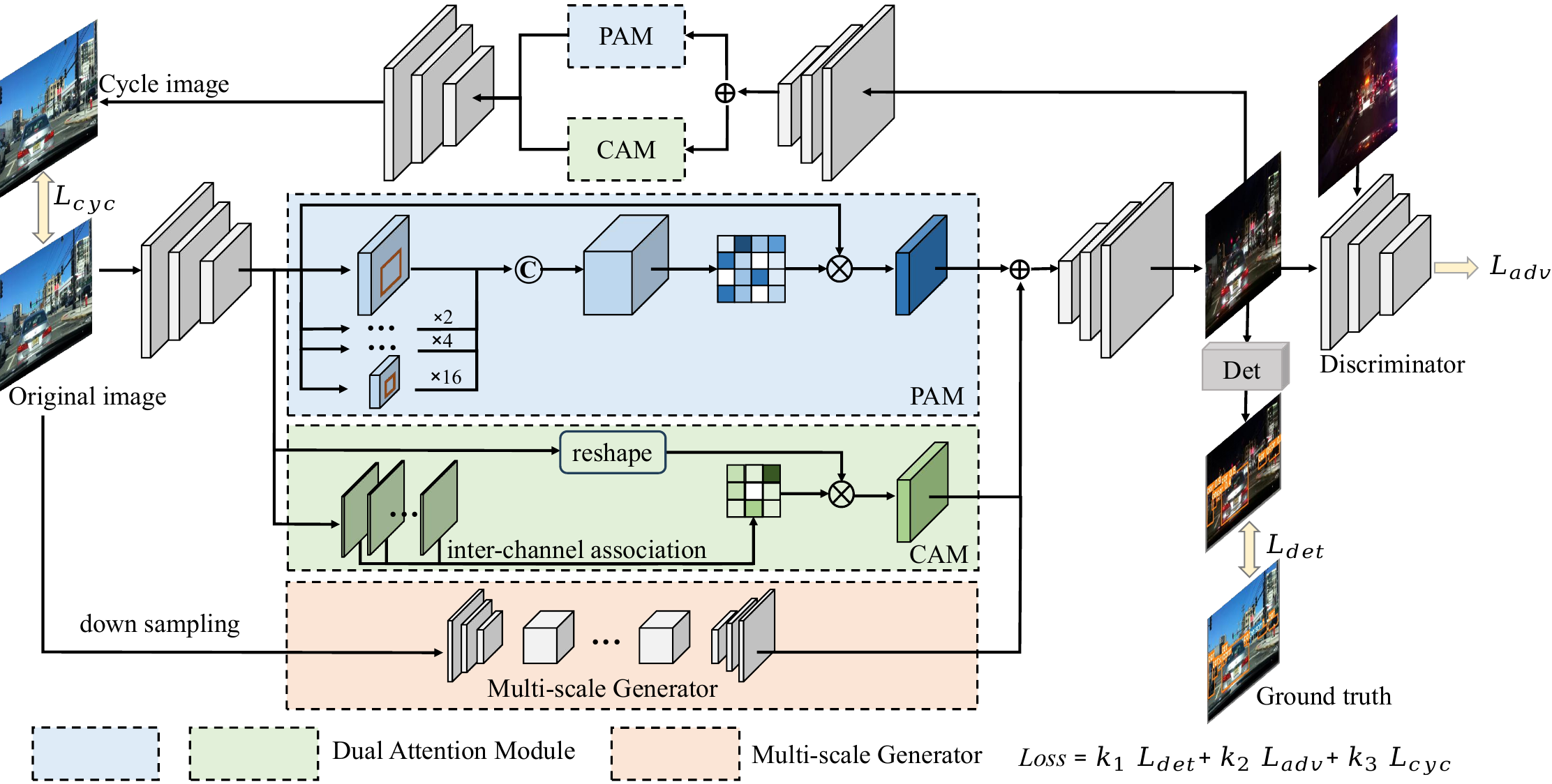} \vspace{-3pt}
  \caption{The framework of our SUSTechGAN contains dual attention modules, multi-scale generators, and the components of the loss function. The dual attention module contains a position attention module (PAM, see \autoref{fig:pam} for the detailed architecture) and a channel attention module (CAM, see \autoref{fig:cam} for the detailed architecture). {\small $Loss$}, {\small $L_{det}$}, {\small $L_{adv}$}, and {\small $L_{cyc}$} represent loss function, detection loss, adversarial loss, and consistency loss respectively}.  \label{fig:network} \vspace{-5pt}
\end{figure*}

Although many studies investigate GAN-based image generation for object detection of autonomous driving \cite{kim2021drivegan,li2021weather,lin2020gan}, there is a lack of studies that investigate this task under adverse conditions.
In our experiments, we applied the well-known GANs (CycleGAN \cite{CycleGAN2017}, UNIT \cite{liu2017unsupervised}, MUNIT \cite{huang2018multimodal}) to generate driving images in adverse conditions of rain and night, as shown in \autoref{fig:rainnight}.  
However, the results (as shown in \autoref{fig:rainnight} and \autoref{tab:statistic}) show that image generation by these existing GANs for driving scenes generally meets the specific issues:
\begin{enumerate}[noitemsep,leftmargin=*,topsep=0pt]
    \item The generated images show weak local semantic features (e.g., vehicles, traffic signs) for object detection.
    For example, vehicles in the generated images of rain and night are generally blurred and even approximately disappeared to imitate the global features of rain and night. 
    Such generated images hardly improve object detection of autonomous driving in adverse conditions.
    \item The generated images show weak global semantic features of adverse conditions. 
    The input images of GANs are generally cropped and resized for small-size images (e.g., 360*360) which hardly cover the global semantic features since images in autonomous driving are generally big-size resolutions such as 1980*1080.
    \item The existing GAN-based image generation methods in autonomous driving have not investigated object detection in adverse conditions.
    Specifically, the existing GANs generally consider the common components of adversarial loss and cycle consistency loss without detection loss 
    for image generation of adverse conditions.
\end{enumerate}



In this paper, we propose a novel framework, SUSTechGAN, to generate driving images for improving object detection of autonomous driving under adverse conditions.
The framework of SUSTechGAN is shown in \autoref{fig:network}.
We conducted ablation studies to investigate the effects of dual attention modules and multi-scale generators in SUSTechGAN.
The results show that both of them significantly contribute to image generation for improving object detection in autonomous driving under adverse conditions.
Furthermore, we compare SUSTechGAN with the well-known GANs for image generation on a customized BDD100K (BDD100k-adv), a new dataset (AllRain), and an adverse condition dataset (ACDC \cite{sakaridis2021acdc}).
We retrain the well-known YOLOv5 by adding generated images into the training dataset.
The results show that generated images by SUSTechGAN significantly improve the retrained YOLOv5 for object detection of driving scenes under adverse conditions, as shown in \autoref{fig:rainnight}.



The main contributions of our GAN-based image generation for improving object detection of autonomous driving in adverse conditions are summarized as:
\begin{enumerate} [noitemsep,leftmargin=*,topsep=0pt]
    \item We design a dual attention module of the position attention module and the channel attention module in SUSTechGAN to improve the local semantic feature extraction for generating driving images in adverse conditions such as rain and night.
    This method solves the issue that the local semantic features (e.g., vehicles) in the generated images are blurred and even approximately disappeared, and improves the object detection of autonomous driving.
    \item We design multi-scale generators in SUSTechGAN to consider various scale features (e.g., big-size generator for global features and small-size generator for local features) for generating high-quality images with clear global semantic features.
    \item We propose a novel loss function with an extra detection loss
    to guide image generation for improving object detection of autonomous driving in adverse conditions.
\end{enumerate}


\vspace{-6pt}
\section{Related Work}
\label{sec:related_work}
\vspace{-1pt}

Here, we review the related studies, including: 1) datasets, 2) generative AI for image generation, 3) GAN-based image generation, and 4) GAN-based driving image generation.

\vspace{-8pt}
\subsection{Datasets}

Several studies \cite{liu2024survey} have discussed autonomous driving datasets.
Although the existing driving datasets generally contain images in various conditions, they contain a few images of real scenes under adverse conditions such as heavy rain weather. 
Furthermore, these images have not been classified for different adverse conditions and can not fully cover various degrees of adverse conditions.
For example, KITTI \cite{geiger2012we} collected driving images in normal weather.
NuScenes \cite{caesar2020nuscenes} contain driving images under moderate weather conditions and lack driving images under various adverse conditions such as light rain, moderate rain, and heavy rain.
Waymo \cite{sun2020scalability} consists of 1150 scenes of each 20 seconds that are collected across suburban and urban and from different times of the day, including day, night, and dawn, rather than different adverse weather conditions.
Cityscapes \cite{cordts2016cityscapes} consists of diverse urban street scenes at varying times of the year.
Although Wang et al. \cite{wang2023effect} considered three weather conditions of rain, fog, and snow, they used synthetic datasets of Foggy Cityscapes, Rain Cityscapes, and Snow Cityscapes to simulate real scenes.
The dataset DENSE contains events and depth maps, which are used to train models for dense monocular depth prediction \cite{hidalgo2020learning}.
However, DENSE is recorded in the CARLA simulator rather than in real driving scenes.
The adverse conditions dataset ACDC contains 4006 images of four adverse conditions: fog, nighttime, rain, and snow \cite{sakaridis2021acdc}. 
Although ACDC covers four adverse weather conditions, it contains only limited degrees of adverse weather conditions since the images were recorded over several days in Switzerland.
In addition, another dataset CADC collected driving scenes during winter weather conditions in Canada, which is basically clear or snowy weather \cite{pitropov2021canadian}.
BDD100k \cite{yu2018bdd100k} contains many driving images in rainy weather, while these images were collected during light rain or after rain which can not fully cover various rain conditions of driving for imitating various rain.
These datasets are hardly used to train GANs to generate high-quality driving images for imitating various degrees of real adverse conditions.
The datasets should contain real driving scenes under diverse degrees of adverse weather conditions.

\vspace{-8pt}
\subsection{Generative AI for Image Generation}
\vspace{-3pt}

Several frameworks have been explored in deep generative modeling, including Variational Autoencoders (VAEs) \cite{vahdat2020nvae}, GANs \cite{CycleGAN2017,liu2017unsupervised,huang2018multimodal} and diffusion models \cite{xu2023diffscene,li2023drivingdiffusion,pronovost2023scenario}.
Arash et al. \cite{vahdat2020nvae} proposed the well-known Nouveau VAE, a deep hierarchical VAE built for image generation using depth-wise separable convolutions and batch normalization.
VAEs generate diverse images quickly, but they generally lack quality.
Xu et al. \cite{xu2023diffscene} proposed a diffusion-based framework, DiffScene, to generate safety-critical scenarios for autonomous vehicle evaluation. 
Li et al. \cite{li2023drivingdiffusion} propose a spatial-temporal consistent diffusion framework, DrivingDiffusion, to generate realistic multi-view videos.
Ethan et al. \cite{pronovost2023scenario} propose a latent diffusion-based architecture for generating traffic scenarios that enable controllable scenario generation. 
Although diffusion-based models could perform outstanding image generation for various tasks, their inference process generally takes a lot of computing.
Zhao et al. \cite{zhao2024exploring} explored these three generative AI methods to create realistic datasets.
The experimental results show that GAN-based methods are adept at generating high-quality images when provided with manually annotated labels for better generalization and stability in driving data synthesis.
A recent remarkable study \cite{berrada2024unlocking} shows that although diffusion models show outstanding performance, GAN-based methods could still perform state-of-the-art performance for image generation.
Specifically, GAN-based methods can surpass recent diffusion models while requiring two orders of magnitude less compute for inference.
While a GAN only needs one forward pass to generate an image, a diffusion model requires several iterative denoising steps, resulting in slower inference \cite{berrada2024unlocking}.

\vspace{-8pt}
\subsection{GAN-based Image Generation} 
\vspace{-3pt}

Many studies have proposed to generate samples from the source domain to the target domain.
Zhu et al. \cite{CycleGAN2017} proposed the well-known CycleGAN to generate images, which possible distortion and corresponding reversible distortion might occur for image generation with cycle loss and brings negative effects.
UNIT \cite{liu2017unsupervised} is another well-known GAN for image generation, which proposed a shared latent space of source and target domains and shared weights between generators.
MUNIT \cite{huang2018multimodal} used the disentangled representation to improve the diversity of generated images.
Recently, Seokbeom et al. \cite{song2023shunit} proposed SHUNIT to generate a new style by harmonizing the target domain style retrieved from a class memory and a source image style.
Although these studies investigated GAN-based image generation,
they have not investigated image generation for driving scenes, particularly under adverse conditions.


Image generation methods in autonomous driving contain traditional physical model-based methods, a combination of computer graphics and physical-based methods, and GAN-based methods.
Traditional physical model-based techniques can be used to generate images of different weather conditions.
Sakaridis et al. \cite{sakaridis2018semantic} build an optical end-to-end model based on stereo matching and depth information to generate foggy images.
Halder et al. \cite{halder2019physics} proposed a physical rendering pipeline to generate rain images by using a physical particle simulator to estimate the position of raindrops and the illumination of rain.
However, these traditional methods generally need empirical physical models that make it hard to simulate real scenes with high definition. 
Many studies produce new scenes by combining high-fidelity computer graphics and physical-based modeling techniques in autonomous driving such as OPV2V \cite{xu2022opv2v}. 
However, the gap between generated images and real images leads to low usability of generated images for autonomous driving, particularly in adverse conditions.

\vspace{-8pt}
\subsection{GAN-based Driving Image Generation}
\vspace{-3pt}

Many studies investigated the effects of adverse weather conditions on object detection.
Wang et al. \cite{wang2023effect} investigated camera data degradation models including light level, adverse weather, and internal sensor noises that are compared for the accuracy of a panoptic segmentation.
Wang et al. \cite{wang2022performance} discussed the effects of adverse illumination and weather conditions on the performance and challenges of 3D object detection for autonomous driving.
However, accurate and robust perception under various adverse weather conditions is still challenging in autonomous driving.

GAN-based methods could generate new samples to augment training datasets under limited data and improve the perception of autonomous driving in adverse conditions.
DriveGAN \cite{kim2021drivegan} is a controllable neural simulator to learn the latent space of images by sampling the style information of images such as weather.
WeatherGAN \cite{li2021weather} changes correlated image regions by learning cues of various weather conditions to generate images.
Lin et al. \cite{lin2020gan} proposed a structure-aware GAN that focuses on a specific task of day-to-night image style transfer rather than image generation for adverse conditions.
Interestingly, Vinicius et al. \cite{arruda2019cross} investigated cross-domain car detection from day to night by using unsupervised
image-to-image translation, while focusing on improving car detection under night conditions.
Although these studies have investigated GAN-based methods for image generation even driving image generation under adverse conditions, there is a lack of studies that focus on image generation for object detection of autonomous driving under adverse conditions.

\vspace{-8pt}
\section{Methodology}
\label{sec:methodology}
\vspace{-3pt}

We customize dual attention modules and multi-scale generators to design a novel framework, SUSTechGAN, to generate driving images under adverse weather conditions. 
In this section, we describe the proposed SUSTechGAN in detail, including dual attention modules of the position attention module and the channel attention module, multi-scale generators, and the novel loss function.
The framework of SUSTechGAN is shown in \autoref{fig:network}.

We define image domains of different conditions (e.g., sunny and rainy images) as source domain {\small $\mathcal{X}$} and target domain {\small $\mathcal{Y}$}.
The samples in source domain $x\in$ {\small $\mathcal{X}$} follow the distribution {\small $\mathcal{P}_x$}, and samples in target domain $y\in$ {\small $\mathcal{Y}$} follow the distribution {\small $\mathcal{P}_y$}. 
For image generation in normal-to-adverse conditions, GANs aim to learn two mapping functions between domains, G:x→y and F:y→x. 

\vspace{-6pt}
\subsection{Dual Attention module}
\vspace{-2pt}

In our preliminary experiments, we applied the well-known CycleGAN \cite{CycleGAN2017}, UNIT \cite{liu2017unsupervised}, and MUNIT \cite{huang2018multimodal} to generate driving images in adverse conditions.
However, we noticed that generated images show weak local semantic features, which are hardly used to improve object detection.
For example, the key objects such as vehicles and traffic signs in generated images under rain weather conditions are generally blurred and even approximately disappeared. 
In this work, we design dual attention modules to improve the semantic features of objects in the generated images.
The dual attention module contains a position attention module (PAM) and channel attention module (CAM), as shown in \autoref{fig:pam} and \autoref{fig:cam} respectively.
Specifically, our dual attention module transforms the output features by PAM and CAM through two 1*1 convolutions respectively, and then sums the feature matrices to fuse the features from PAM and CAM.

\subsubsection{Position Attention Module}

\begin{figure*}[!ht]  \centering \small
  \includegraphics[width=0.95\linewidth,trim={0 328 3 0},clip]{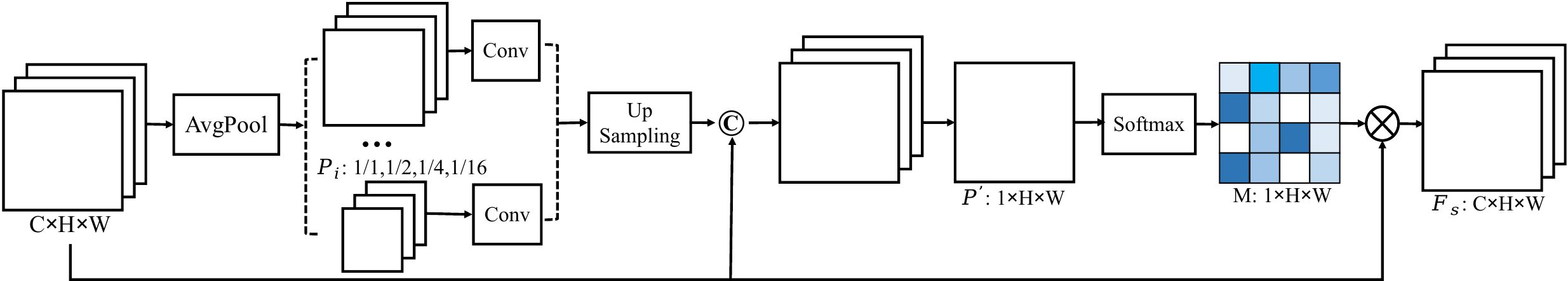} \vspace{-3pt}
  \caption{The framework of Position Attention Module (PAM) in our dual attention module.}  \label{fig:pam}
\end{figure*}

We apply PAM to SUSTechGAN to improve local semantic feature extraction in big-size driving images.
Specifically, pixels in an image are associated with each other from a semantic feature by PAM. 
The aggregation operation in PAM extracts the position association of pixels and fuses the association into a semantic feature.
The PAM finally produces the position-weighted feature that is the same size as the original feature of input images.
The framework of PAM is shown in \autoref{fig:pam}.

In PAM, the features are operated by 1/1, 1/2, 1/4, and 1/16 down-sampling, which generates 4 different size features.
The feature sizes can be controlled by the size of pooling kernels. 
The four size features are upsampled by bilinear interpolation to recover the same size as the original features and merge various scale features channel by channel.
The low-dimensional pooling convolution result is upsampled by bilinear interpolation to restore the feature map of the same size as the original feature.
The feature is updated as {\small $\mathcal{P}^{'} \in \mathbb{R}^{1 \times H \times W}$}.
The attention matrix {\small $\mathcal{M} \in \mathbb{R}^{1 \times H \times W}$} can be normalized by softmax:
\begin{equation} \centering \small
    \mathcal{M} = \sigma \mathcal{P}^{'}, \text{where} ~\mathcal{P}^{'} =concat(\mathcal{P}_1, \mathcal{P}_2, \mathcal{P}_3, \mathcal{P}_4)
\end{equation}
where $\sigma$ and $concat$ represent normalization function sigmoid and concatenation operation in channels.
Finally, the feature {\small $\mathcal{F}_s\in\mathbb{R}^{C\times H\times W}$} can be updated by the operation operation between the attention matrix and the original features as:
\begin{equation} \centering \small
    \mathcal{F}_s = \mathcal{M} \otimes \mathcal{F}
\end{equation}
where $\otimes$ represents the multiplication operation of matrices.

\subsubsection{Channel Attention Module}

Autonomous driving scenes generally have various semantic features and high complexity. 
The generator in the exiting GANs generally extracts features by convolution layers, which hardly extracts and expresses various and complex semantic features of driving scenes.
We apply channel attention modules to extract the association of features in various channels that improve feature expression.

\begin{figure} [!ht] \centering
  \includegraphics[width=0.98\linewidth,trim={285 142 163 129},clip]{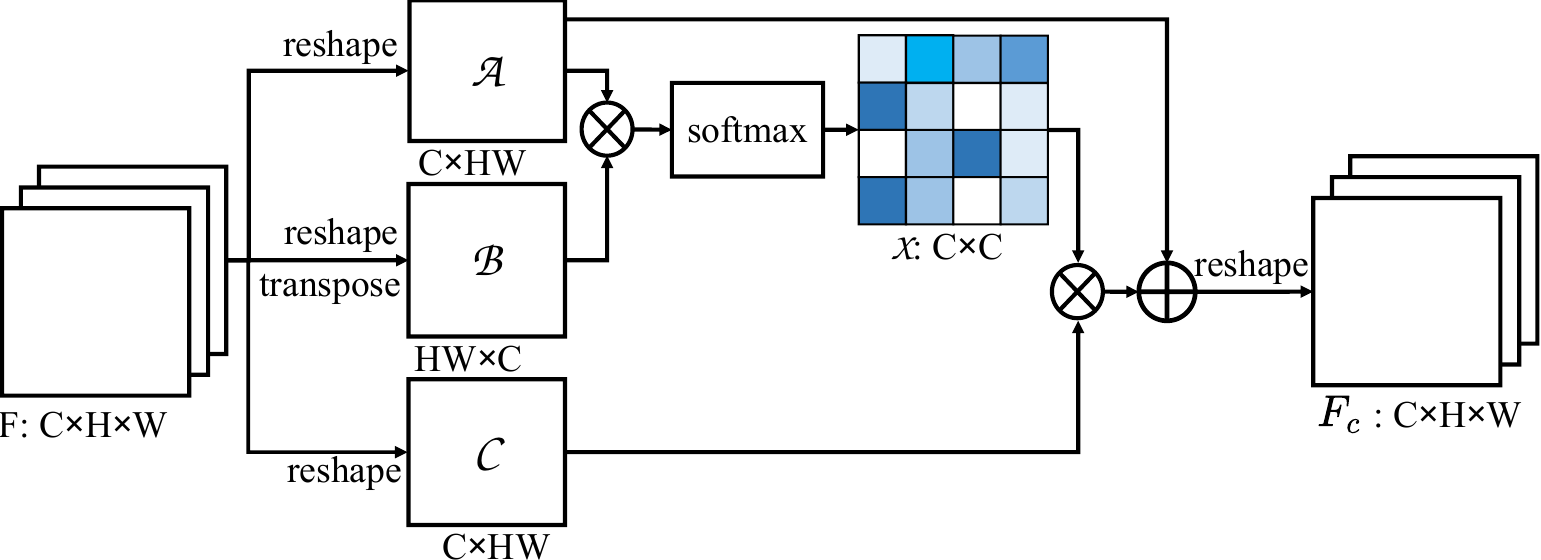} \vspace{-5pt}
  \caption{The framework of Channel Attention Module (CAM).}  \label{fig:cam}
\end{figure}

In autonomous driving, scenes generally contain rich and complex semantic features due to various challenges such as blurring and occlusion caused by complex and adverse conditions like various weather and light effects. 
The features in different channels can be associated with different types of semantic features.
The CAM in the dual attention module aggregates the associated features from different channels to improve feature extraction and expression.
The framework of CAM is shown in \autoref{fig:cam}.
\vspace{-5pt}

Specifically, the CAM reshape the original features {\small $\mathcal{F} \in \mathbb{R}^{C\times H\times W}$} to {\small $\mathcal{A} \in \mathbb{A}^{C\times HW}$}, {\small $\mathcal{B} \in \mathbb{R}^{C\times HW}$} and {\small $\mathcal{C}\in \mathbb{R}^{C\times HW}$}.
The product of {\small $\mathcal{A}$} and transpose matrices of {\small $\mathcal{B}$} is normalized by softmax operation and produces the attention matrix {\small $\mathcal{X} \in \mathbb{R}^{C\times C}$} where each element $x$ represents the association between different channels.
\begin{equation} \centering \small
        x =  \exp (\mathcal{A}·\mathcal{B}^T)/{\sum}^{\mathcal{C}} \exp(\mathcal{A}·\mathcal{B}^T)
\end{equation}
The attention matrix {\small $\mathcal{X}$} multiplies {\small $\mathcal{C}$$\in$$\mathbb{R}^{C\times N}$} (where {\small $N$$=$$H$$\times$$W$}), and reshape the product result to {\small $\mathbb{R}^{C\times H\times W}$}.
The feature with attention weights is added to the original feature and thus the channel attention {\small $\mathcal{F}_c \in \mathbb{R}^{C\times H\times W}$} can be calculated by:
\begin{equation} \centering \small
        \mathcal{F}_c =  \mathcal{X} \otimes \mathcal{C} + \mathcal{A}
\end{equation}
Finally, the improved features can be calculated by adding the outputs of the dual attention module and the multi-scale generators, as the process shown in \autoref{fig:network}.

\vspace{-8pt}
\subsection{Multi-scale Generator}
\vspace{-2pt}
The real driving scenes contain local semantic features like vehicles and global semantic features like adverse weather conditions.
Therefore, the generated images should contain clear local features and global features for adverse conditions. 
However, the generators in the existing GANs hardly generate driving images for considering both local and global features since the input images of GANs are generally cropped and resized for the same small scale (e.g., 360*360) which hardly cover the global semantic features of big-size resolutions images such as 1980*1080 in autonomous driving.
In our preliminary experiments, we tried to use bigger scale inputs such as 720x720 in the well-known GANs, which significantly increases the training time with a negligible improvement for trained object detection.
In this work, we develop the multi-scale generators to SUSTechGAN for combining local and global semantic features in the generated images.

\begin{figure}[!ht]  \centering \small
  \includegraphics[width=0.98\linewidth,trim={212 198 300 111},clip]{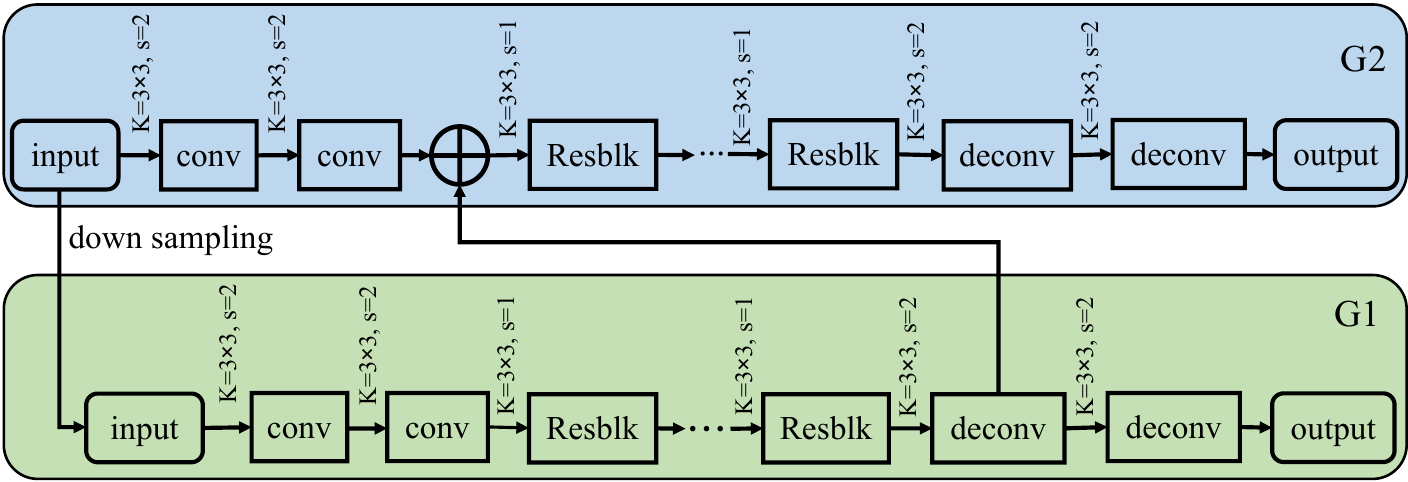} \vspace{-3pt}
  \caption{The framework of multi-scale generators in our SUSTechGAN, where $k$ represents the kernel size and $s$ represents the convolution stride.
  } \label{fig:generator}
\end{figure}

We design SUSTechGAN with two embedded generators {\small $\mathcal{G}_1$}, {\small $\mathcal{G}_2$} of different scales, as shown in \autoref{fig:generator}.
Specifically, the generator {\small $\mathcal{G}_1$} resizes the input feature by $1/4$ followed by the three operations of down-sampling, residual block, and deconvolution to generate an image with a $1/4$ size of input images.
The generator {\small $\mathcal{G}_2$} contains convolution for feature extraction, residual block, and deconvolution.
\vspace{-5pt}
To fuse the features from different generators, the generator {\small $\mathcal{G}_1$} is embedded into the generator {\small $\mathcal{G}_2$}.
The framework of multi-scale generators and the connections between generators are shown in \autoref{fig:generator}.
In particular, the feature after deconvolution in the generator {\small $\mathcal{G}_1$} is added to the feature after two convolutions in the generator {\small $\mathcal{G}_2$}, which integrates multiple the two generators. 
The lower-layer generators are added to the higher-layer generators to fuse local semantic features such as vehicles and global features such as adverse weather conditions.

\vspace{-8pt}
\subsection{Loss Function}
\vspace{-2pt}

We propose a novel loss function with an extra detection loss, that is usually not considered in the existing GAN-based image generation, to guide image generation for improving object detection of autonomous driving in adverse conditions.
We proposed a detection loss by considering the performance of the pre-retrained object detection on generated images, which is compared to the ground truth.
In this work, we mainly consider a detection loss ({\small $\mathcal{L}_{det}$}) and the common loss of adversarial loss ({\small $\mathcal{L}_{adv}$}) and cycle consistency loss ({\small $\mathcal{L}_{cyc}$}) to guide the training process of SUSTechGAN by:
\begin{equation} \centering
    \mathcal{L}_{total}=k_1 \mathcal{L}_{det} + k_2 \mathcal{L}_{adv} + k_3 \mathcal{L}_{cyc} 
\end{equation}
where the weights $k_1 = 0.8$, $k_2 = 1$, and $k_3 = 10$ are naive-tuned and considered the empirical values to balance detection loss, adversarial loss, and cycle consistency loss, respectively.
 
For detection loss $\mathcal{L}_{det}$, we apply the pre-trained YOLOv5 on BDD100k to detect generated images and consider the performance as the detection loss. 
Referring to the loss function in YOLOv5, we consider the detection loss with localization loss, classification loss, and confidence loss as:
\begin{equation} \centering
    \mathcal{L}_{det}=a \mathcal{L}_{CIoU} + b \mathcal{L}_{cls} + c \mathcal{L}_{conf} 
\end{equation}
where $a=0.4$, $b=0.3$, $c=0.3$ refers to YOLOv5. 
The localization loss considers the intersection over union representing the ratio of intersection area and union area of predicted box and ground truth.
The classification loss considers the predicted possibility of the current class.
Last, the confidence loss indicates the confidence of predicted bounding boxes.

For adversarial loss $\mathcal{L}_{adv}$, the total adversarial loss is:
\begin{equation} \centering
    \mathcal{L}_{adv}= \mathcal{L}_{adv}(\mathcal{G},\mathcal{D}_{\mathcal{Y}})+\mathcal{L}_{adv}(\mathcal{F},\mathcal{D}_{\mathcal{X}})
\end{equation}
where $\mathcal{L}_{adv}(\mathcal{G},\mathcal{D}_\mathcal{Y})$ and $\mathcal{L}_{adv}(\mathcal{F},\mathcal{D}_{\mathcal{X}})$ represent cycle loss of normal-to-adverse {\small $\mathcal{G}: \mathcal{X} \to \mathcal{Y}$} and inverse cycle loss of adverse-to-normal {\small $\mathcal{F}: \mathcal{Y} \to \mathcal{X}$}.

For cycle consistency loss $\mathcal{L}_{cyc}$, each generated sample {\small $\mathcal{G}_(x)$} from source domain {\small $\mathcal{X}$} should be brought back to the original image, i.e., {\small $x \to \mathcal{G}(x) \to \mathcal{F}(\mathcal{G}(x)) \approx x$}.
The cycle consistency loss can be expressed as: 
\begin{equation} \centering \small
    \mathcal{L}_{cyc}(\mathcal{G},\mathcal{F})=\mathbb{E}_x[{||\mathcal{F}(\mathcal{G}(x))-x||}_1]+\mathbb{E}_y[{||\mathcal{G}(\mathcal{F}(x))-y||}_1]
\end{equation}

\vspace{-8pt}
\section{Experiments}
\label{sec:experiments}
\vspace{-3pt}

In this section, we address the experiments in detail,
including datasets, evaluation, and experimental setup.

\vspace{-11pt}
\subsection{Datasets} 
\label{subsec:datasets}
\vspace{-2pt}

In this work, we conduct experiments on the three datasets: 1) a customized BDD100k (BDD100k-adv) where images are classified into sunny, rainy, and night weather conditions, 2) a novel dataset (AllRain) where driving images were collected from YouTube videos under various degrees of rainy conditions, 3) a well-known adverse condition dataset, ACDC.
Here, we address these three datasets in detail.
Note that the dataset AllRain has not collected nighttime driving images since BDD100k-adv contains enough nighttime driving images.
The specifications of the datasets are shown in \autoref{tab:dataset}. 
\begin{table} [!ht] \centering \small  
\setlength\tabcolsep{1pt} \renewcommand{\arraystretch}{0.9}
    \begin{tabular}{lclc} \toprule 
    Dataset & Sunny (train/test) & Rainy (train/test)  &  Night (train/test) \\  \midrule
    BDD100k-adv & 2000(1600/400) & 1342(1000/342) & 2500 (2000/500) \\
    AllRain & 2827(2314/513) & 2121(1736/385) & --- \\ 
    ACDC & --- & 1000(500/500) & 1006 (506/500) \\ \bottomrule
    \end{tabular} \vspace{-4pt}
    \caption{The specifications of the adverse weather datasets.} 
    \label{tab:dataset}
\end{table} 

\subsubsection{BDD100k-adv}
The original BDD100k contains 100,000 video clips from multiple cities and under various conditions.
For each video, a keyframe with a resolution of 1280x720 is selected with detailed annotations such as the bounding box of different objects. 
The images in BDD100k are classified into six various conditions, including clear, overcast, and rainy. 
In this work, we selected 1600/400 (train/test) sunny images, 1000/342 (train/test) rainy images, and 2000/500 (train/test) night images to compose the dataset, named BDD100k-adv.

\subsubsection{AllRain (A novel dataset)}

The existing datasets can not fully cover various degrees of adverse conditions.
We designed a new customized dataset, named \textit{AllRain}. 
Specifically, we reviewed many driving videos from YouTube and selected some videos that contain driving on highways and urban streets under sunny and various degrees of rain such as light rain, moderate rain, and heavy rain.
We extracted images from the videos and used LabelImg to annotate these images with three common categories: pedestrian, car, and truck.
The images are with a large resolution of 1920×1080.
Finally, the dataset AllRain contains 2314/513 (train/test) sunny images and 1736/385 (train/test) rainy images, as shown in \autoref{tab:dataset}.

\subsubsection{ACDC}
\label{subsubsec:acdc}

ACDC is a driving dataset with 4006 images that are recorded from several days under four adverse conditions: fog,
nighttime, rain, and snow on highways and in rural regions of Switzerland. 
The images are categorized into 1000 foggy, 1006
nighttime, 1000 rainy, and 1000 snowy images.

\vspace{-6pt}
\subsection{Evaluation}
\label{subsec:evaluation}
\vspace{-2pt}

\subsubsection{Image Quality Metric} 

In general, image quality can be assessed from various metrics such as FID (Frechet Inception Distance) for similarity and KID (Kernel Inception Distance) for diversity, which have been widely used to assess the visual quality of generated images by GAN-based methods.
Specifically, FID is generally used to evaluate the similarity between generated and real images.
A lower FID score indicates the higher-quality generated images that are distributed similarly to the real images.
KID is a metric similar to the FID that measures the distance between the real image distribution and the generated image distribution to evaluate the image quality.
FID assumes that the features extracted by Inception obey the normal distribution, which is a biased estimate. 
KID performs unbiased estimation based on Maximum Mean Discrepancy, and the calculation includes an unbiased estimate of a cubic kernel.
In this work, we adopt FID and KID to assess the generated image quality in terms of the similarity between generated images and real images and the diversity of generated images.

\subsubsection{Object Detection Metrics}
We retrain the well-known YOLOv5 by adding generated images to test the improvement of object detection in adverse conditions and evaluate the retrained YOLOv5 by using mean average precision (mAP). 
Notice that each generated image has the same ground truth as the original real image for calculating mAP. 

\vspace{-8pt}
\subsection{Experimental Setup} 
\vspace{-2pt}


In this work, we designed experiments to compare our SUSTechGAN with the well-known GANs, CycleGAN, UNIT, and MUNIT by following their original instructions.
In addition, we demonstrate the effects of dual attention modules and multi-scale generators in SUSTechGAN by ablation studies. 

Specifically, we implement and train SUSTechGAN, CycleGAN, UNIT, and MUNIT with PyTorch on an NVIDIA Tesla V100 GPU system. 
SUSTechGAN, CycleGAN, and UNIT took approximately 15.5 hours of average computation time, while MUNIT took approximately 18.7 hours.
In the training stage, we resize images to a width of 1080 followed by a random cropping of size 360×360 for the dataset AllRain.
For the dataset BDD100k-adv, the training stage randomly crops the images to a size of 720×720. 
In addition, the Adam solver 
was employed for optimization, where hyper-parameters ${\beta}_1$ and ${\beta}_2$ were set to 0.9 and 0.999, respectively.
The batch size was set to 1, and the initial learning rate was set to 0.0002, as the default values in CycleGAN.

\begin{figure*}[!b] \centering \small
    \begin{subfigure}[t]{.24\textwidth}
    \includegraphics[width=\textwidth]{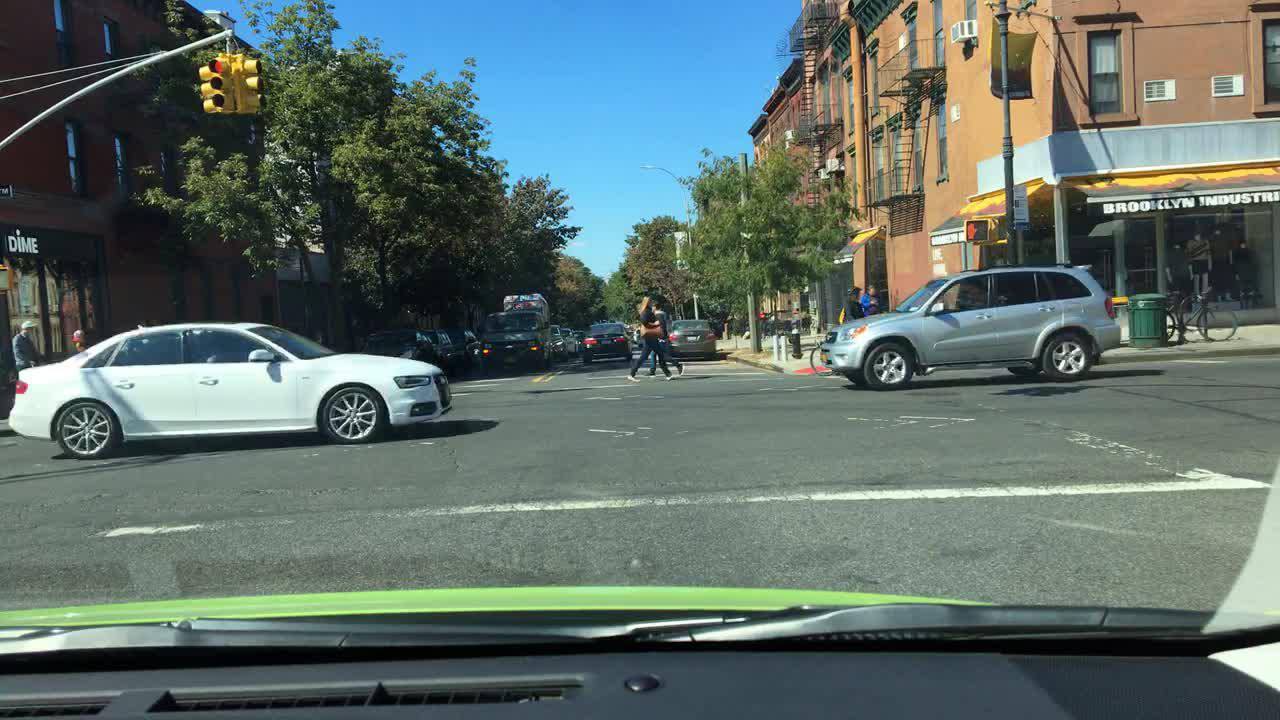} \vspace{-15pt}
    \caption{Original image} \label{fig:car1}
    \end{subfigure}
    \begin{subfigure}  [t]{.24\textwidth}
    \includegraphics[width=\textwidth]{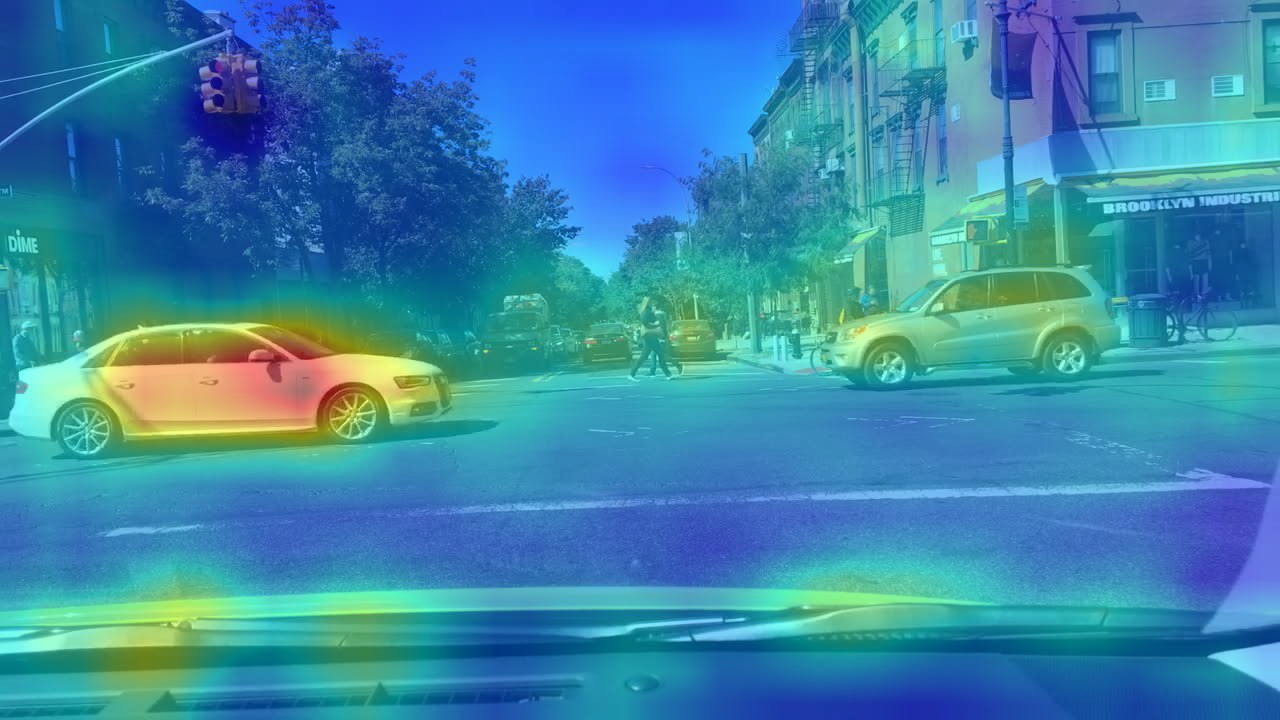} \vspace{-15pt}
    \caption{Position attention}\label{fig:pam1}
    \end{subfigure}
    \begin{subfigure}  [t]{.24\textwidth}
    \includegraphics[width=\textwidth]{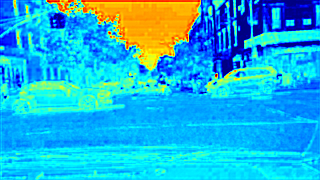} \vspace{-15pt}
    \caption{Channel (\#6) attention}\label{fig:cam1}
    \end{subfigure}
    \begin{subfigure}  [t]{.24\textwidth}
    \includegraphics[width=\textwidth]{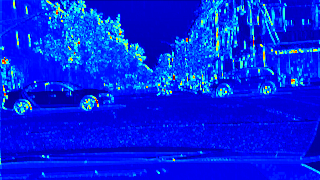} \vspace{-15pt}
    \caption{Channel (\#13) attention}\label{fig:cam1-2}
    \end{subfigure}  \vspace{-1pt}
    \caption{The feature results after position attention module and channel attention module.} \label{fig:attention}
\end{figure*}

\vspace{-6pt}
\section{Results}
\label{sec:results}
\vspace{-1pt}

In this section, we present the qualitative and quantitative results to demonstrate our method.
Specifically, we present the results of the ablation study and analyze the key parameter (generator scale) in SUSTechGAN.
Furthermore, we present the results of the retrained YOLOv5 on the various combinations of generated images and real images under adverse conditions, e.g., rainy and night conditions.
Finally, we compare SUSTechGAN with the well-known GANs, CycleGAN, UNIT and MUNIT.

\vspace{-6pt}
\subsection{Ablation Study and Parameter Tuning}
\label{subsec:ablation}
\vspace{-2pt}


\subsubsection{Ablation Study}

We analyze the effects of 1) the PAM and CAM in the dual attention module and 2) the detection loss and present the results in \autoref{tab:pam_cam}. 
\begin{table}[!ht]  \centering \small
\setlength\tabcolsep{3pt} \renewcommand{\arraystretch}{0.8}
    \begin{tabular}{l|cccc} \toprule
    Methods & PAM & CAM & FID$\downarrow$ & mAP$\uparrow$ \\  \midrule
    Baseline & \ding{55} & \ding{55} & 76.5 & 0.382\\ \midrule
    $\mathcal{L}_{total}-\mathcal{L}_{det}$ & \ding{55} & \ding{55} & 79.7(4.2\%$\uparrow$) & 0.361(5.5\%$\downarrow$) \\ \midrule
    \multirow{3}{*}{\makecell[l]{Dual \\ attention \\ module}} & \checkmark & \ding{55} & 69.2 ($9.5\% \downarrow$) &  0.418 ($9.4\% \uparrow$) \\
    & \ding{55} & \checkmark & 70.7 ($7.6\% \downarrow$)  &  0.415 ($8.6\% \uparrow$) \\
    & \cellcolor{gray!30} \checkmark & \cellcolor{gray!30} \checkmark &  \cellcolor{gray!30}\textbf{67.3} ($12.0\% \downarrow$) & \cellcolor{gray!30}\textbf{0.422} ($10.5\% \uparrow$) \\ \bottomrule
    \end{tabular} \vspace{-3pt}
    \caption{The results of ablation study on BDD100k-adv.
    } \vspace{-1pt} \label{tab:pam_cam}
\end{table} 
The results show that both PAM and CAM contribute to generating driving images with lower FID and higher mAP.
Specifically, the generator in SUSTechGAN with PAM performs a result with $9.5\%$ lower FID and $9.4\%$ higher mAP than the baseline.
Similarly, the generator with CAM performs $7.6\%$ lower FID and $8.6\%$ higher mAP than the baseline.
Finally, the generator with the combination of PAM and CAM performs $12.0\%$ lower FID and $10.5\%$ higher mAP than the baseline.
In addition, we conduct an ablation study to investigate the effect of the detection loss. 
The results show the trained generator without detection loss performs $4.2\%$ higher FID and $5.5\%$ lower mAP than the baseline. 
Finally, we investigate the contribution of the multi-scale generator and dual attention module in SUSTechGAN, as the results shown in \autoref{tab:fid}.
The results show that both the multi-scale generator and dual attention module contribute to the lower FID and KID of generated images.
Last, we show the feature results after PAM and various channel CAMs in \autoref{fig:attention}.
The images in the second column show the attention features of key objects after PAM which the positions in the same object are associated with an attention feature.
The images in the third and fourth columns show the features after channels 6 and 13 attention where the significantly different features from different channels are extracted. 
For example, channel 6 extracts the sky features. 
With PAM and CAM, the generator in SUSTechGAN benefits from the extracted features to generate high-quality driving images for improving object detection of autonomous driving under adverse conditions.

\subsubsection{Parameter Tuning}

\begin{figure}[!ht]  \centering \small
  \includegraphics[width=0.98\linewidth,trim={5 3 37 31},clip]{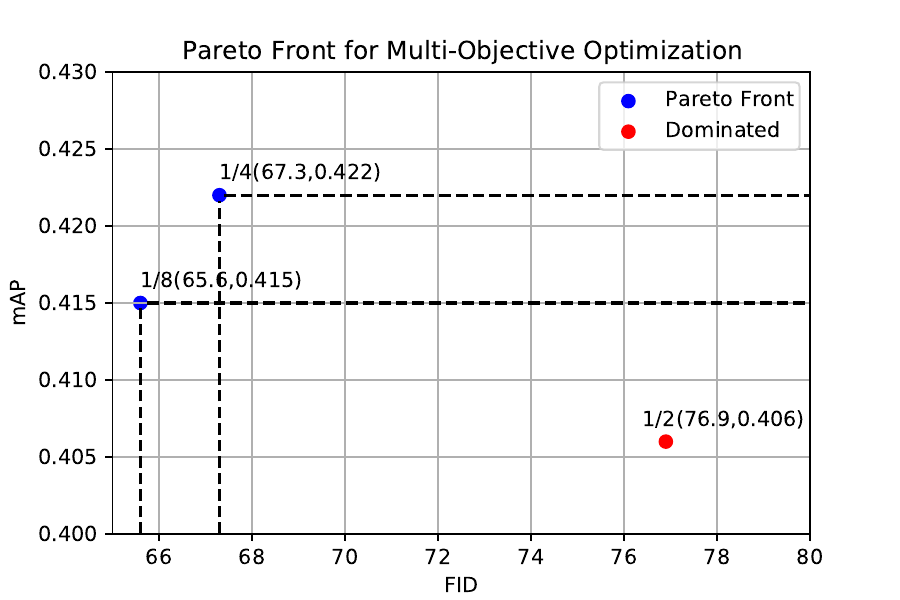} \vspace{-5pt}
  \caption{The Pareto front results of various down-sampling scales in the generator of SUSTechGAN. The FID and mAP are the average values on the rain images of driving.}  \vspace{-3pt} \label{fig:parameter}
\end{figure}

The down-sampling scale in the generator of SUSTechGAN is an important parameter. 
We analyze their performance with different scale values of down-sampling.
In this work, we evaluate the down-sampling scale from 1) FID between the generated images and real images and 2) the average mAP of retrained YOLOv5 by the combined training dataset of generated images and real images.
The down-sampling scale of the generator with lower FID and higher mAP is better.
Therefore, we choose a down-sampling scale value by considering the Pareto front of the multi-objective FID and mAP, as shown in \autoref{fig:parameter}.
The results show that the down-sampling scale values of 1/4 (67.3,0.422) and 1/8 (65.6,0.415) dominate 1/2 (76.9,0.406).
Both 1/4 and 1/8 are the Pareto front in terms of FID and mAP. 
Here, we chose 1/4 as the down-sampling scale since it has a higher mAP with a similar FID of 1/8.
\vspace{-5pt}

\vspace{-6pt}
\subsection{Object Detection}
\label{subsec:objectrecognition}

We apply the generated driving images under adverse conditions to retrain YOLOv5 and test the retrained YOLOv5. 
We apply SUSTechGAN to generate driving images for rain and night conditions and evaluate the distance between generated images and real images by FID and KID.
\def\rot{\rotatebox}
\begin{table} [!ht] \centering \footnotesize  
\setlength\tabcolsep{1pt} \renewcommand{\arraystretch}{0.9}
    \begin{tabular}{l l l c c l l} \toprule 
    \multicolumn{2}{c}{\rot{30}{Methods}} & \rot{30}{\makecell[c]{AllRain\\(Rainy)}} & \rot{25}{\scriptsize\makecell[c]{BDD100k-adv\\(Rainy)}} & \rot{25}{\scriptsize\makecell[c]{BDD100k-adv\\(Night)}} & \rot{30}{\makecell[c]{ACDC\\(Rainy)}} & \rot{30}{\makecell[c]{ACDC\\(Night)}} \\ \midrule
    \multirow{3}{*}{Base} & {\scriptsize CycleGAN} & 79.8/0.12 & 89.2/0.09 & 71.7/0.05 & 85.1/0.10 & 69.9/0.04 \\
    & UNIT & 81.3/0.12 & 85.6/0.08 & \textbf{61.2}/0.04 & 83.4/0.09 & 60.7/0.04  \\
    & MUNIT & 92.5/0.14 & 107.1/0.09 & 83.9/0.06 & 101.7/0.11 & 80.5/0.05 \\ \midrule
    \multirow{2}{*}{Ours} & mul. & 70.1/0.12 & 76.5/0.08 & 66.5/0.04 & 74.3/0.09 & 63.1/0.03 \\
    & \cellcolor{gray!30}att.+mul. & \cellcolor{gray!30}\textbf{55.2}/\cellcolor{gray!30}\textbf{0.11} & \cellcolor{gray!30}\textbf{67.3}/\cellcolor{gray!30}\textbf{0.07} & \cellcolor{gray!30}64.8/\cellcolor{gray!30}\textbf{0.02} & \cellcolor{gray!30}\textbf{61.6/0.08} & \cellcolor{gray!30}\textbf{59.8/0.02} \\ \bottomrule
    \end{tabular}  \vspace{-3pt}
    \caption{The FID$\downarrow$ / KID$\downarrow$ between the generated images by GANs and the real driving images. 
    } \label{tab:fid}
\end{table} 
We compare SUSTechGAN with the well-known CycleGAN, UNIT, and MUNIT, as shown in \autoref{tab:fid}.
The results show that our SUSTechGAN generates driving images on rain with lower FID (i.e., closer between generated images and real images) than the other GANs on all three datasets.
The generated driving images at night by both UNIT and our proposed GAN have a low FID and KID, while the generated driving images at night by UNIT have a lower FID and KID.

\begin{figure} [!ht] \centering \small \vspace{-3pt}
  \includegraphics[width=0.97\linewidth,trim={155 45 175 53},clip]{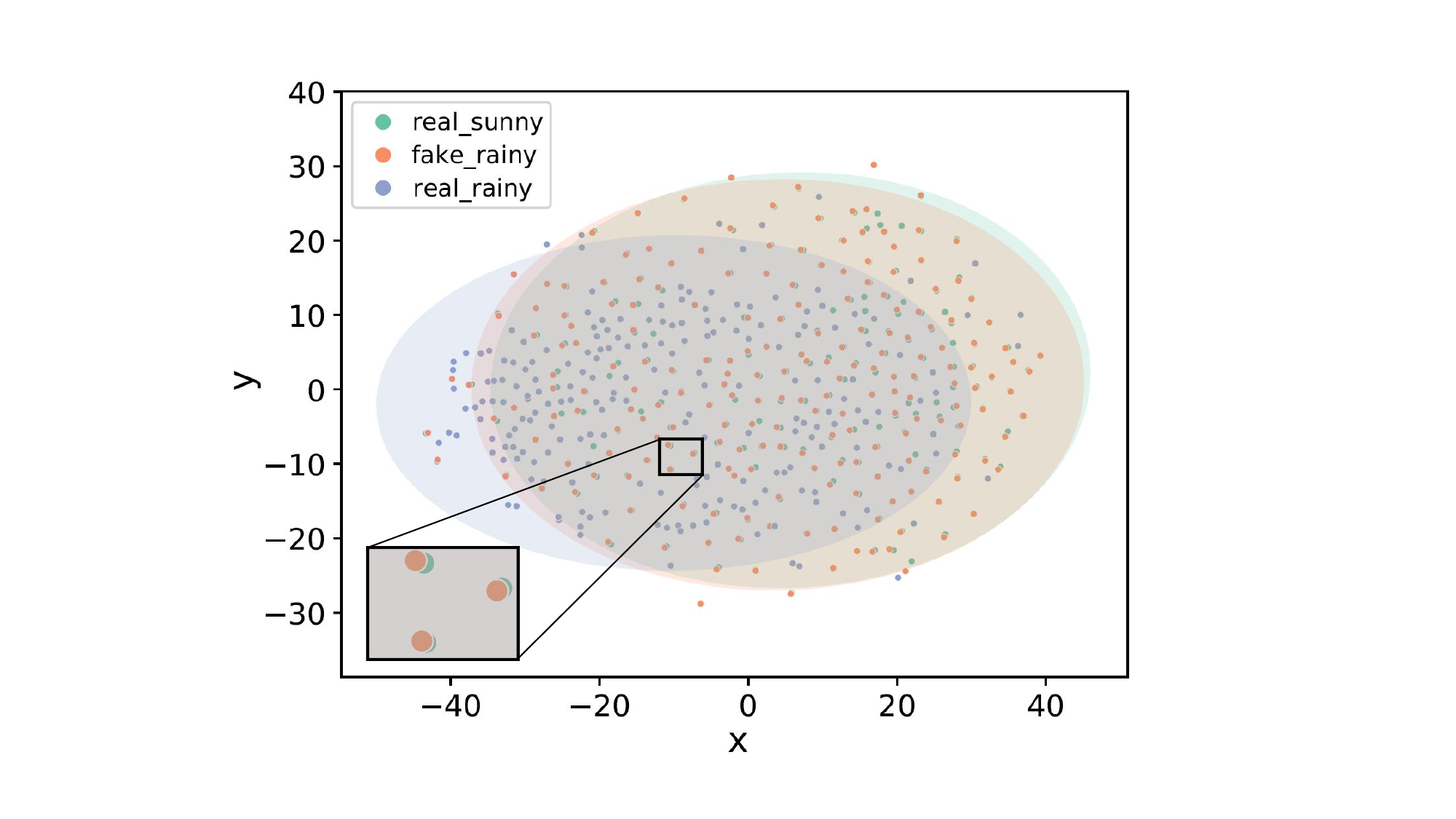}
  \vspace{-5pt}
  \caption{The visualization of the semantic features of dimensionality reduction for real sunny and rainy images of driving, and the generated rainy driving images.}  \label{fig:feature}
\end{figure}

Furthermore, we analyse the domains of generated images and real images by dimensionality reduction for their convolution features of YOLOv5.
The convolution features are dimensionality reduced into two-dimensional samples by using the well-known t-SNE 
and visualized in \autoref{fig:feature}.
All the two-dimensional samples constitute a feature matrix $\mathcal{T}$.
The mean $(x,y)$ of the feature matrix $\mathcal{T}$ in the $x$ and $y$ direction is the center point of the ellipse.
We calculate the eigenvalues and eigenvectors of the covariance matrix of the feature matrix $\mathcal{T}$ in the $x$ and $y$ directions. 
The direction of the eigenvector is the orientation of the ellipse, and the eigenvalues are the radius of the ellipse in the $x$ and $y$ directions.
The results show that the key objects in the generated rain images and the real sunny images have similar convolutional features, which indicates that the generated rainy images by SUSTechGAN retain the local semantic features (e.g., vehicles).
The semantic feature domain of the generated and real rainy images are largely overlapped.
The semantic feature domains of the generated rain images and the real rain images are closed, which indicates the generated rain images can be used to improve object detection for autonomous driving.

We compare SUSTechGAN with CycleGAN, UNIT, and MUNIT by adding their generated images under rain and night conditions into the training set of YOLOv5.
In this work, we retrain YOLOv5 by the three training settings: 
\begin{enumerate} [noitemsep,leftmargin=*,topsep=0pt]
\item Real daytime images. Only real daytime images (including sunny images) are used in the training dataset. 
\item Real rainy or night images added. 200 real images under rain or night conditions are added to the training set. 
\item Generated images added. 
200 generated images by our SUSTechGAN and other GANs are added to the training.
\end{enumerate}

The retrained YOLOv5 is tested on real driving images of rain and night conditions.
The (mean) average precision of the retrained YOLOv5 under three training settings is shown in \autoref{tab:detection}.
For rain conditions, the retrained YOLOv5 under training setting 2) performed a best mAP of 0.457.
However, the retrained YOLOv5 on the adding generated images by CycleGAN, UNIT, and MUNIT performed much lower mAP than the adding real images of driving.
The retrained YOLOv5 under training setting 3) by adding the generated rain images by SUSTechGAN performed a 0.422 mAP which is a 47.4\% improvement of the training setting 2) with fully real rain images. 
For night conditions, the retrained YOLOv5 under training data setting 3) with adding generated night images by SUSTechGAN performed a 0.469 mAP which is a 19.6\% improvement of the training setting 2) with fully real night images and outperforms the method with a 10.5\% improvement in reference \cite{arruda2019cross}. 
The result indicates that the generated driving images by SUSTechGAN can improve object detection of autonomous driving in adverse conditions.

\begin{table}[!ht] \centering \small
\setlength\tabcolsep{2pt} \renewcommand{\arraystretch}{1.0}
    \begin{tabular}{l l c c c c} \toprule
    Scenes & Data & Pedestrian & Car & Truck & mAP \\ \midrule
    \multirow{6}{*}{Rainy} & \cellcolor{gray!30}1) Sunny(Real) & \cellcolor{gray!30}0.286 & \cellcolor{gray!30}0.361 & \cellcolor{gray!30}0.283 & \cellcolor{gray!30}0.310(100\%) \\ \cline{2-6}
    & 2) +Rainy(Real) & 0.467 & 0.473 & 0.431 & 0.457 \\ \cline{2-6}
    & 3) +CycleGAN & 0.360 & 0.458 & 0.343 & 0.357 \\
    & \quad +UNIT & 0.340 & 0.452 & 0.333 & 0.375 \\
    & \quad +MUNIT & 0.314 & 0.417 & 0.322 & 0.351 \\
    & \cellcolor{gray!30}\quad \textbf{+Ours} & \cellcolor{gray!30} \textbf{0.409} & \cellcolor{gray!30} \textbf{0.470} & \cellcolor{gray!30} \textbf{0.387} & \cellcolor{gray!30} \textbf{0.422(47.4\%$\uparrow$)} \\ \midrule
    \multirow{6}{*}{Night} & \cellcolor{gray!30}1) Day(Real) & \cellcolor{gray!30}0.306 & \cellcolor{gray!30}0.448 & \cellcolor{gray!30}0.421 & \cellcolor{gray!30}0.392(100\%) \\ \cline{2-6}
    & 2) +Night(Real) & 0.489 & 0.542 & 0.528 & 0.52 \\ \cline{2-6}
    & 3) +CycleGAN & 0.325 & 0.461 & 0.445 & 0.410 \\
    & \quad +UNIT & 0.338 & 0.477 & 0.459 & 0.425 \\
    & \quad +MUNIT & 0.363 & 0.470 & 0.471 & 0.435 \\
    & \cellcolor{gray!30}\quad \textbf{+Ours} & \cellcolor{gray!30}\textbf{0.408} & \cellcolor{gray!30}\textbf{0.516} & \cellcolor{gray!30}\textbf{0.482} & \cellcolor{gray!30}\textbf{0.469(19.6\%$\uparrow$)} \\ \bottomrule
    \end{tabular} \vspace{-2pt}
    \caption{The performance of the retrained YOLOv5 by adding 200 generated images to the training dataset.}
    \label{tab:detection}
\end{table} 
Furthermore, we retrain YOLOv5 with the addition of various numbers of generated images by SUSTechGAN and CycleGAN and compare their performance in terms of mAP over an increasing number of generated images, as shown in \autoref{fig:addition}.
The results show that the retrained YOLOv5 by the addition of generated rain or night images of driving by our SUSTechGAN outperforms that of CycleGAN.
For the adverse conditions of driving in rain and night conditions, the retrained YOLOv5 performs an increasing mAP when the generated images are increasingly added to the training settings until they approach a converged mAP.

\sidecaptionvpos{figure}{c}
\begin{SCfigure*}[10][!ht] \centering
\begin{tikzpicture}[scale=0.95]
    \begin{axis}[
    ymin = 0.3,
    ymax = 0.55,
    symbolic x coords={+100,+200,+500,+1000,+1300,+1500,+2000},
    height=0.7\linewidth,
    width=0.9\linewidth,
    grid=major,
    xlabel={Nunmber of added images},
    ylabel={mAP},
    xlabel style={yshift=5pt},
    ylabel style={yshift=-13pt},
    legend style={
    legend pos=south east, }]
    \addplot +[line width=1pt] coordinates {(+100,0.337) (+200,0.422)  (+500,0.448)  (+1000,0.481) (+1300,0.525)  (+1500,0.530)  (+2000,0.527)} 
    node at (0,70) {0.337}
    node[pos=0.17,above] {0.422}
    node[pos=0.33,above] {0.448}
    node[pos=0.5,above] {0.481}
    node[pos=0.67,above] {0.525}
    node[pos=0.83,above] {0.530}
    node[pos=1,above] {0.527};
    \addplot +[line width=1pt] coordinates {(+100,0.342) (+200,0.387)  (+500,0.425)  (+1000,0.462) (+1300,0.506)  (+1500,0.518)  (+2000,0.509)}
    node at (0,20) {0.342}
    node[pos=0.17,below] {0.387}
    node[pos=0.33,below] {0.425}
    node[pos=0.5,below] {0.462}
    node[pos=0.67,below] {0.506}
    node[pos=0.83,below] {0.518}
    node[pos=1,below] {0.509};
    \legend{Ours,CycleGAN}
    \end{axis}
\end{tikzpicture} \vspace{-3pt}
\begin{tikzpicture}[scale=0.95]
    \begin{axis}[
    ymin = 0.3,
    ymax = 0.55,
    symbolic x coords={+100,+200,+500,+1000,+1300,+1500,+2000},
    height=0.7\linewidth,
    width=0.9\linewidth,
    grid=major,
    xlabel={Nunmber of added images},
    ylabel={mAP},
    xlabel style={yshift=5pt},
    ylabel style={yshift=-13pt},
    legend style={
    legend pos=south east, }]
    \addplot +[line width=1pt] coordinates {(+100,0.432) (+200,0.469)  (+500,0.481)  (+1000,0.494) (+1300,0.516)  (+1500,0.507)  (+2000,0.510)} 
    node at (0,155) {0.432}
    node[pos=0.17,above] {0.469}
    node[pos=0.33,above] {0.481}
    node[pos=0.5,above] {0.494}
    node[pos=0.67,above] {0.516}
    node[pos=0.83,above] {0.507}
    node[pos=1,above] {0.510};
    \addplot +[line width=1pt] coordinates {(+100,0.396) (+200,0.410)  (+500,0.463)  (+1000,0.469) (+1300,0.447)  (+1500,0.451)  (+2000,0.457)}
    node at (0,75) {0.396}
    node[pos=0.17,below] {0.410}
    node[pos=0.34,below] {0.463}
    node[pos=0.5,below] {0.469}
    node[pos=0.67,below] {0.447}
    node[pos=0.83,below] {0.451}
    node[pos=1,below] {0.457};
    \legend{Ours,CycleGAN}
    \end{axis}
\end{tikzpicture} \vspace{-3pt}
\caption{The mAP of retrained YOLOv5 by adding various numbers of generated rain (left) and night (right) images by our SUSTechGAN and CycleGAN to the training dataset.} \label{fig:addition}
\end{SCfigure*}

\begin{table}[!ht] \centering \small
\setlength\tabcolsep{2pt} \renewcommand{\arraystretch}{0.9} 
    \begin{tabular}{l | c c c c >{\columncolor[gray]{0.8}}c} \toprule
    Ground truth & metrics & CycleGAN & UNIT & MUNIT & Ours \\  \midrule
    \multirow{3}{*}{Rainy (2128)} & FN & \textbf{568} & 601 & 663 & \textbf{411}(27.6\%$\downarrow$) \\
    & FP & 226 & 151 & \textbf{102} & \textbf{93}(8.8\%$\downarrow$) \\
    & TP & \textbf{1560} & 1527 & 1465 & \textbf{1717}(10.1\%$\uparrow$) \\ \midrule
    \multirow{3}{*}{Night (3292)} & FN & 1524 & 1323 & \textbf{1250} & \textbf{1057}(7.9\%$\downarrow$) \\
    & FP & 1193 & \textbf{945} & 1074 & \textbf{870}(27.6\%$\downarrow$) \\
    & TP & 1768 & 1969 & \textbf{2042} & \textbf{2235}(9.5\%$\uparrow$) \\  \bottomrule
    \end{tabular} \vspace{-2pt}
    \caption{The statistic results of the pre-trained YOLOv5 by BDD100k tested on 200 generated rain and night images. False Negative (FN), False Positive (FP), True Positive (TP).} \label{tab:statistic} 
\end{table} 

\begin{figure*}[!ht] \centering
    \begin{subfigure}[t]{.195\textwidth}
    \includegraphics[width=\textwidth]{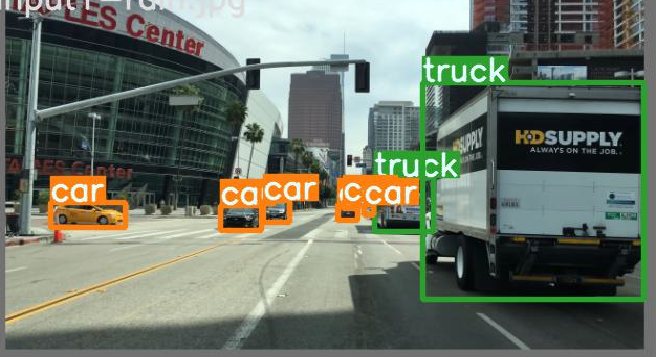}
    \label{fig:input1} \vspace{-10pt} 
    \end{subfigure} \hspace{-5pt}
    \begin{subfigure} [t]{.195\textwidth}
    \includegraphics[width=\textwidth]{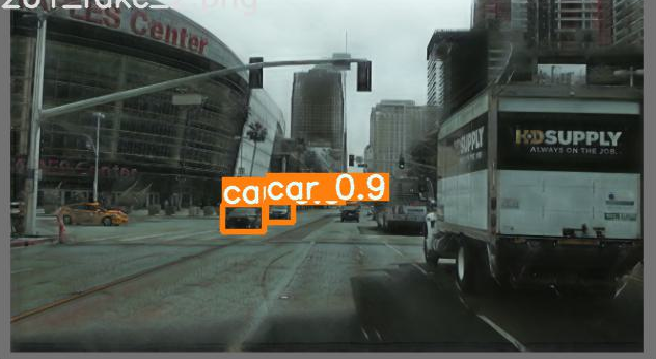}
    \label{fig:cyc1} \vspace{-10pt}
    \end{subfigure} \hspace{-5pt}
    \begin{subfigure} [t]{.195\textwidth}
    \includegraphics[width=\textwidth]{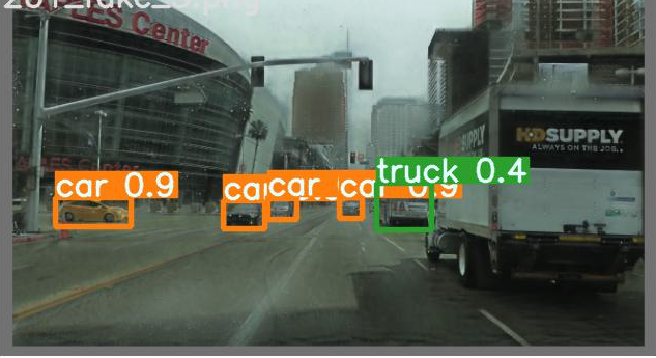}
    \label{fig:unit1} \vspace{-10pt}
    \end{subfigure} \hspace{-5pt}
    \begin{subfigure} [t]{.195\textwidth}
    \includegraphics[width=\textwidth]{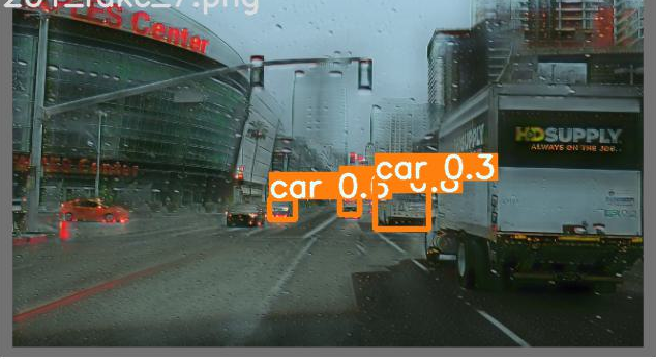}
    \label{fig:munit1} \vspace{-10pt}
    \end{subfigure} \hspace{-5pt}
    \begin{subfigure} [t]{.195\textwidth}
    \includegraphics[width=\textwidth]{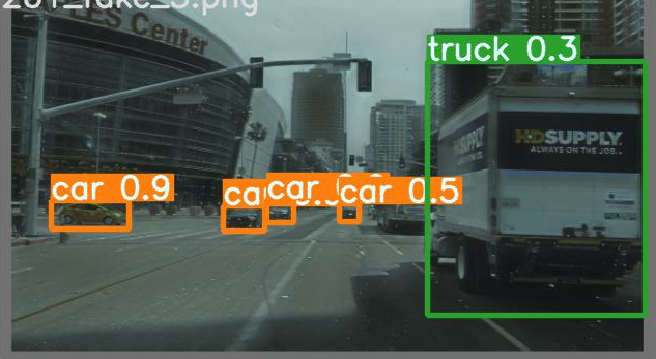}
    \label{fig:ours1} \vspace{-10pt}
    \end{subfigure} \vspace{-10pt}
    \begin{subfigure} [t]{.195\textwidth}
    \includegraphics[width=\textwidth]{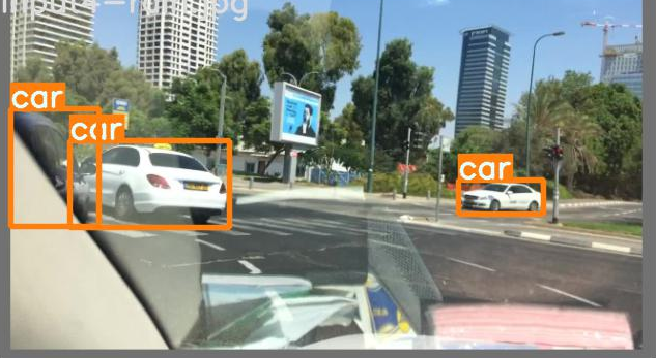}
    \end{subfigure} \hspace{-5pt}
    \begin{subfigure} [t]{.195\textwidth}
    \includegraphics[width=\textwidth]{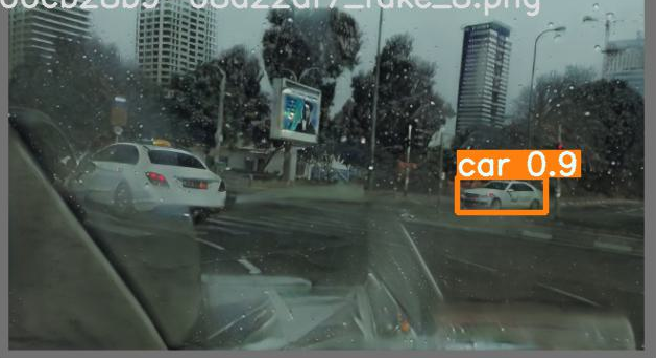}
    \end{subfigure} \hspace{-5pt}
    \begin{subfigure} [t]{.195\textwidth}
    \includegraphics[width=\textwidth]{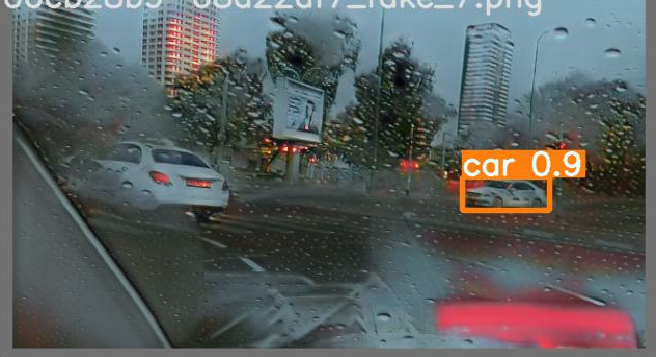}
    \end{subfigure} \hspace{-5pt}
   \begin{subfigure} [t]{.195\textwidth}
    \includegraphics[width=\textwidth]{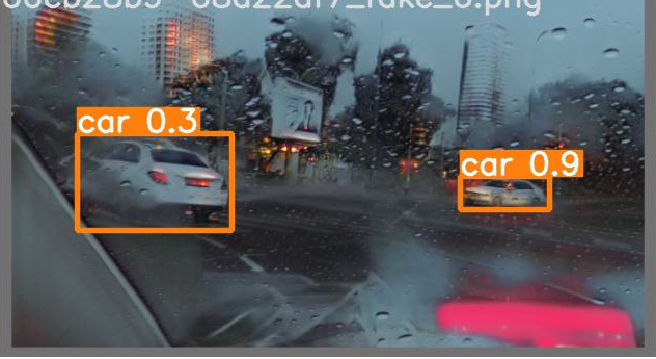}
    \end{subfigure} \hspace{-5pt}
    \begin{subfigure} [t]{.195\textwidth}
    \includegraphics[width=\textwidth]{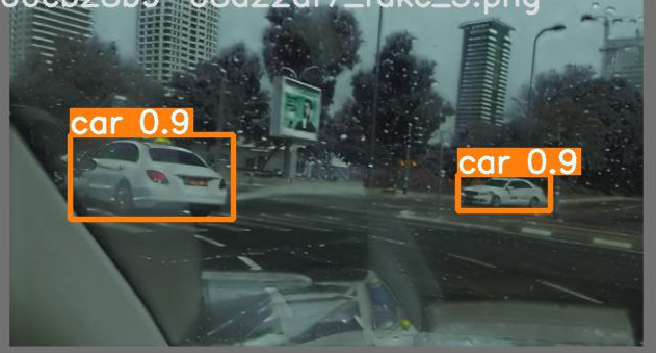}
    \label{fig:ours4rain}
    \end{subfigure} \vspace{-10pt}
    \begin{subfigure}[t]{.195\textwidth}
    \includegraphics[width=\textwidth]{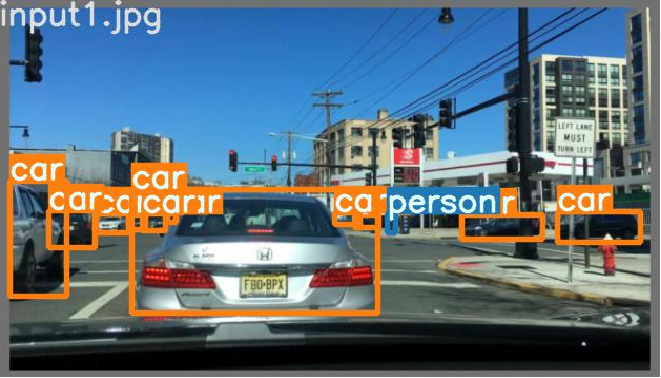}
    \label{fig:input}
    \end{subfigure} \hspace{-5pt}
    \begin{subfigure} [t]{.195\textwidth}
    \includegraphics[width=\textwidth]{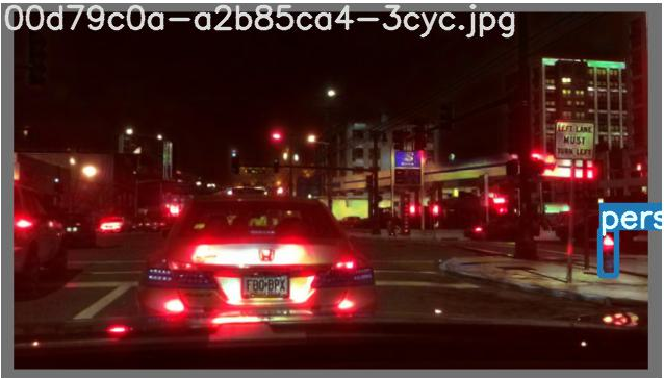}
    \label{fig:cyc}
    \end{subfigure} \hspace{-5pt}
    \begin{subfigure} [t]{.195\textwidth}
    \includegraphics[width=\textwidth]{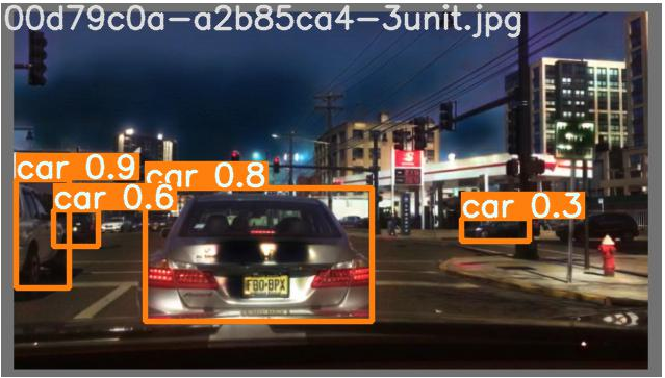}
    \label{fig:unit}
    \end{subfigure} \hspace{-5pt}
    \begin{subfigure} [t]{.195\textwidth}
    \includegraphics[width=\textwidth]{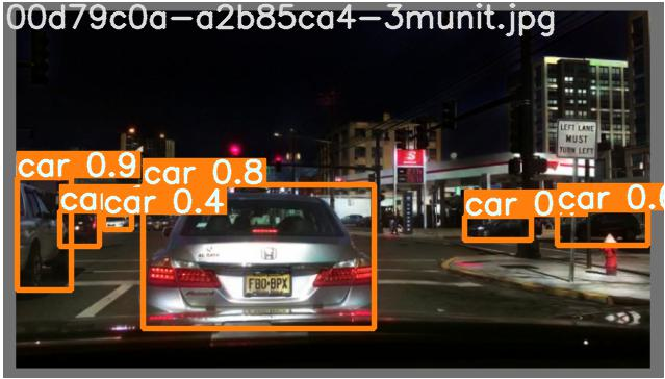}
    \label{fig:munit}
    \end{subfigure} \hspace{-5pt}
    \begin{subfigure} [t]{.195\textwidth}
    \includegraphics[width=\textwidth]{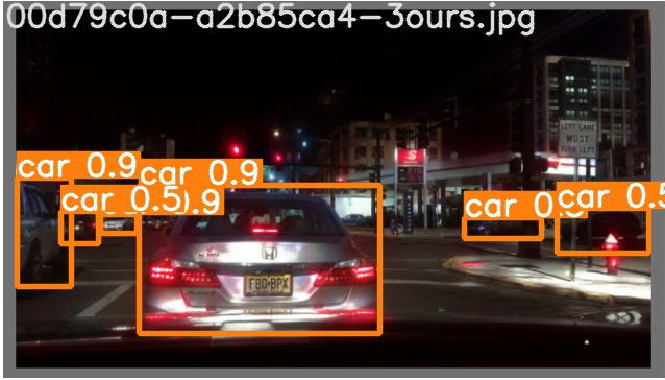}
    \label{fig:ours} 
    \end{subfigure} \vspace{-1pt} \hspace{0.05pt}
    \begin{subfigure} [t]{.195\textwidth}
    \includegraphics[width=\textwidth]{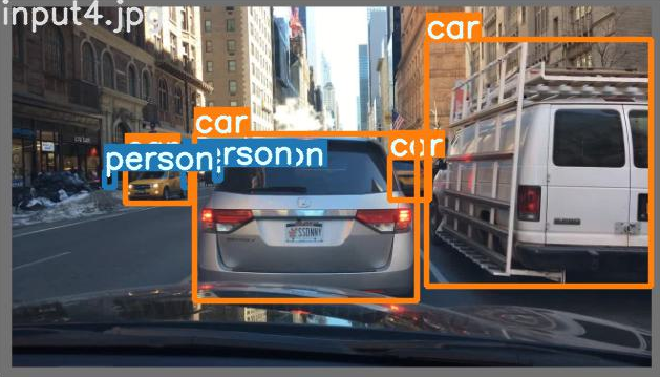} \vspace{-15pt}
    \caption{Ground Truth}
    \end{subfigure} \hspace{-5pt}
    \begin{subfigure} [t]{.196\textwidth}
    \includegraphics[width=\textwidth]{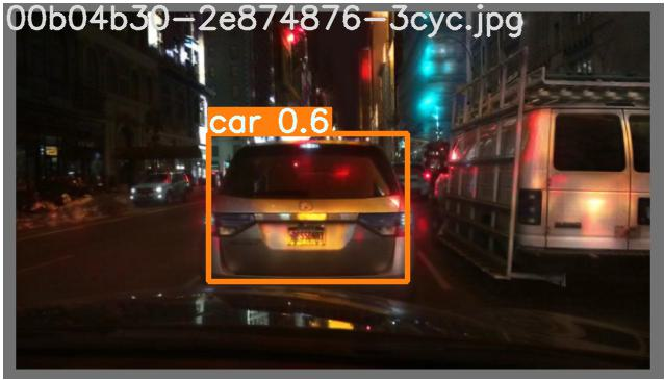} \vspace{-15pt}
    \caption{CycleGAN}
    \end{subfigure} \hspace{-5pt}
    \begin{subfigure} [t]{.196\textwidth}
    \includegraphics[width=\textwidth]{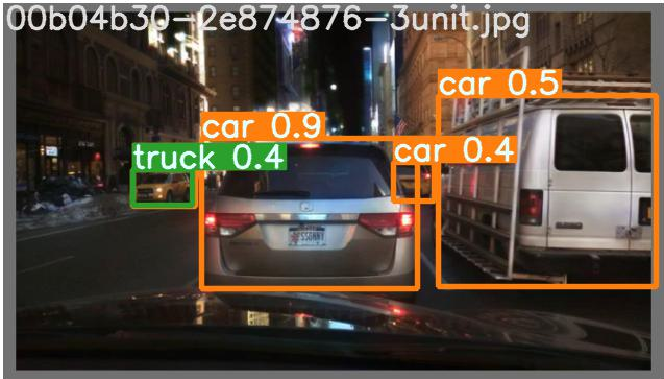} \vspace{-15pt}
    \caption{UNIT}
    \end{subfigure} \hspace{-5pt}
   \begin{subfigure} [t]{.196\textwidth}
    \includegraphics[width=\textwidth]{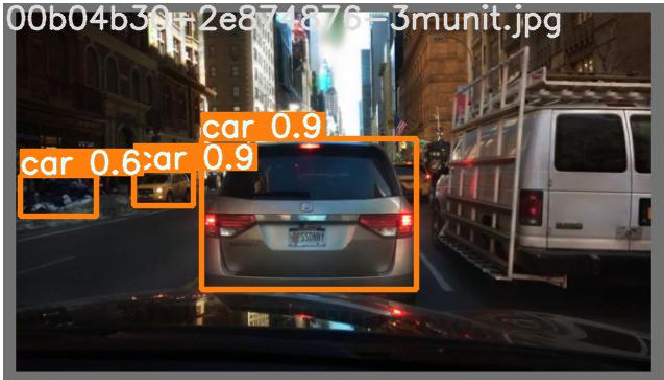} \vspace{-15pt}
    \caption{MUNIT}
    \end{subfigure} \hspace{-4.5pt}
    \begin{subfigure} [t]{.195\textwidth}
    \includegraphics[width=\textwidth]{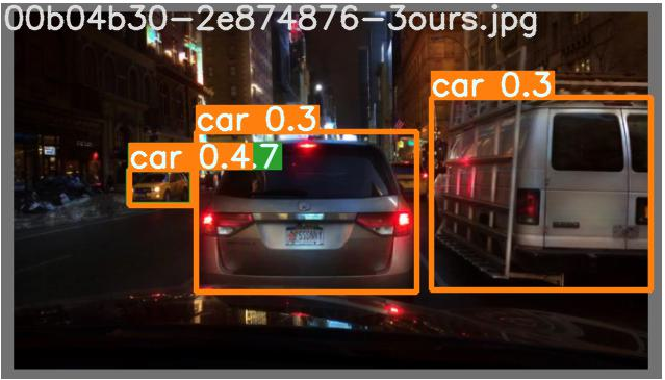} \vspace{-15pt}
    \caption{Ours}
    \label{fig:ours4night}
    \end{subfigure} \vspace{-1pt}
    \caption{The object detection results of the pre-trained YOLOv5 on the generated images by CycleGAN, UNIT, MUNIT, and our SUSTechGAN under the rain (first and second rows) and night (third and fourth rows) conditions. a) shows the real daytime images of four random driving scenes where the bounding boxes are the labeled ground truth.} \label{fig:rainnight}
\end{figure*}

We applied various GANs to generate driving images under rain and night conditions as a testing set and tested the pre-trained YOLOv5 by BDD100k on the generated driving images. 
The statistic results of object detection by the pre-trained YOLOv5 are shown in \autoref{tab:statistic}.
The object detection results of the pre-trained YOLOv5 on the generated rain and night images of four random driving scenes by different GANs are shown in \autoref{fig:rainnight}.
The real daytime images of four random driving scenes are the inputs of the generators of various GANs, where the bounding boxes are the labeled ground truth.
Specifically, several objects in the generated rain images by CycleGAN can not be detected by the pre-trained YOLOv5.
For UNIT and MUNIT in \autoref{fig:rainnight}, several local semantic features are weak in the generated rain images.
For example, the bus, cars, and buildings in the first and second rows are blurred.
Even worse, there are heavy light effects, such as the red color in the generated images, which does not match the real images of driving in the rain.
Due to these defects, the objects in the generated images by UNIT and MUNIT can not be detected correctly.
In contrast, the generated rain images of driving by our SUSTechGAN show proper objects and backgrounds under rain without unreasonable light effects. 
The pre-trained YOLOv5 performed a much better object detection on the generated rain images by SUSTechGAN than the others.
For the night condition in \autoref{fig:rainnight}, the generated images by CycleGAN have many chaotic glare that blurs images and reduces object visibility, e.g., the third row of CycleGAN.
The objects thus can not be detected by the pre-trained YOLOv5.
The generated night images by UNIT and MUNIT have distortion defects such as the bright light of day in the fourth row of UNIT and MUNIT and the black blocks in the third row of UNIT.
Our SUSTechGAN generates better night images of driving without glare and distortion defects than the others, which can be applied to generate driving images for improving object detection of autonomous driving under adverse conditions.

\vspace{-6pt}
\section{Discussion}
\label{sec:discussion}

The experimental results have demonstrated that our SUSTechGAN outperforms the well-known GANs in image generation for improving object detection of autonomous driving in adverse conditions.
However, many open issues could be interesting and need to be discussed in depth.
In addition, some limitations of this work would be challenging and need to be investigated further. 

In SUSTechGAN, we referred to the well-known GANs for the channel parameter settings in the channel attention module, which could be further investigated for improvement.
Furthermore, the effects of different channels have not been analyzed in-depth and need to be investigated.
In \autoref{fig:addition}, we notice that the retrained YOLOv5 performs a significantly increasing mAP with a low initial value when adding numbers of generated rain images into the training dataset.
In contrast, the retrained YOLOv5 has a slow-increasing mAP with a moderate initial value when adding numbers of generated night images into the training dataset.
For this issue, we suppose that rainy scenes are more complex than night scenes, and rainy scenes are more diverse than night scenes.
For example, there are various degrees of rain such as light rain, moderate rain, and heavy rain for rain conditions.
The night scenes could be less diverse and easier to retrain object detection models for a convergent performance than rain scenes.

Furthermore, we discuss the potential limitations of this work.
\textit{1) Structure:} This work focuses on image generation of driving in adverse conditions by applying the dual attention module and multi-scale mechanism for the normal-to-adverse generator.
To reduce the training time, we develop both dual attention modules and multi-scale mechanism to the normal-to-adverse generator but only dual attention modules without the multi-scale mechanism to the adverse-to-normal generator, in which the adverse-to-normal generator is weaker than the normal-to-adverse.
The weaker adverse-to-normal generator may slightly decrease the normal-to-adverse generator, which could be improved. 
\textit{2) Dataset:} We validate the proposed SUSTechGAN on small-scale datasets (BDD100k-adv, AllRain, and ACDC) due to the limited driving images under adverse weather conditions. 
Nonetheless, our SUSTechGAN has been demonstrated to improve object detection of autonomous driving in adverse conditions on these datasets.
\textit{3) Methods:} Both our SUSTechGAN and the well-known GANs generate driving images for adverse conditions by considering global features, which may perform weak local features of specific objects.
For example, the generated night images have various glare that is different from real night images, while our SUSTechGAN with the attention module performs better than the existing well-known GANs.

\vspace{-8pt}
\section{Conclusion}
\label{sec:conclusion}


In this work, we propose a novel SUSTechGAN to generate images for training object detection of autonomous driving in adverse conditions.
We design dual attention modules and multi-scale generators in SUStechGAN to improve the local and global semantic features for various adverse conditions such as rain and night.
We test SUSTechGAN and the well-known GANs to generate driving images in adverse conditions of rain and night and apply the generated images to retrain the well-known YOLOv5. 
Specifically, we add generated images into the training datasets for retraining YOLOv5 and evaluate the improvement of the retrained YOLOv5 for object detection in adverse conditions.
The experimental results demonstrate that the generated driving images by our SUSTechGAN significantly improved the performance of retrained YOLOv5 in rain and night conditions, which outperforms the well-known GANs. 
In the future, we will investigate the open issues and limitations discussed in \autoref{sec:discussion}.
Furthermore, we will apply SUSTechGAN to generate driving images under more adverse conditions such as snow and fog.

\vspace{-6pt}


\vspace{-3pt} 
\bibliographystyle{IEEEtran}
\bibliography{bibliography}

\vspace{-13mm}
\begin{IEEEbiography}
    [{\includegraphics[width=1in,height=1.25in,trim={10 30 15 0}, clip, keepaspectratio]{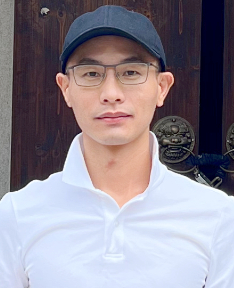}}]
    {Gongjin Lan} is currently a Research Assistant Professor at the Department of Computer Science and Engineering, Southern University of Science and Technology, Shenzhen, China. 
    He received his PhD in Artificial Intelligence at VU University Amsterdam in 2020, published over 20 academic papers in top international journals and conferences, including \textit{IEEE T-HMS, IEEE T-LT, Swarm and Evolutionary Computation, Scientific Reports, PPSN}, and serves as a reviewer for \textit{IEEE T-EVC, Swarm and Evolutionary Computation, Scientific Reports, PLOS ONE, ICRA, IROS, GECCO, CEC}, etc. His research interests include Autonomous Driving, Deep Learning, and AI interdisciplinary. 
\end{IEEEbiography}

\vspace{-13mm}
\begin{IEEEbiography}
    [{\includegraphics[width=1in,height=1.25in,trim={5 35 5 5}, clip, keepaspectratio]{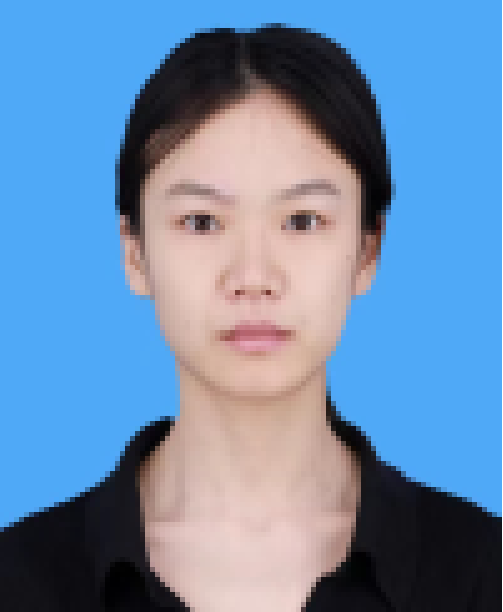}}]
    {Yang Peng} received her M.E. at the Department of Computer Science and Engineering, Southern University of Science and Technology, Shenzhen, China, Dec. 2023. 
    Her research interests include Generative adversarial networks and deep neural networks. 
\end{IEEEbiography}

\vspace{-13mm}
\begin{IEEEbiography}
    [{\includegraphics[width=1in,height=1.25in,trim={0 55 0 65},clip,keepaspectratio]{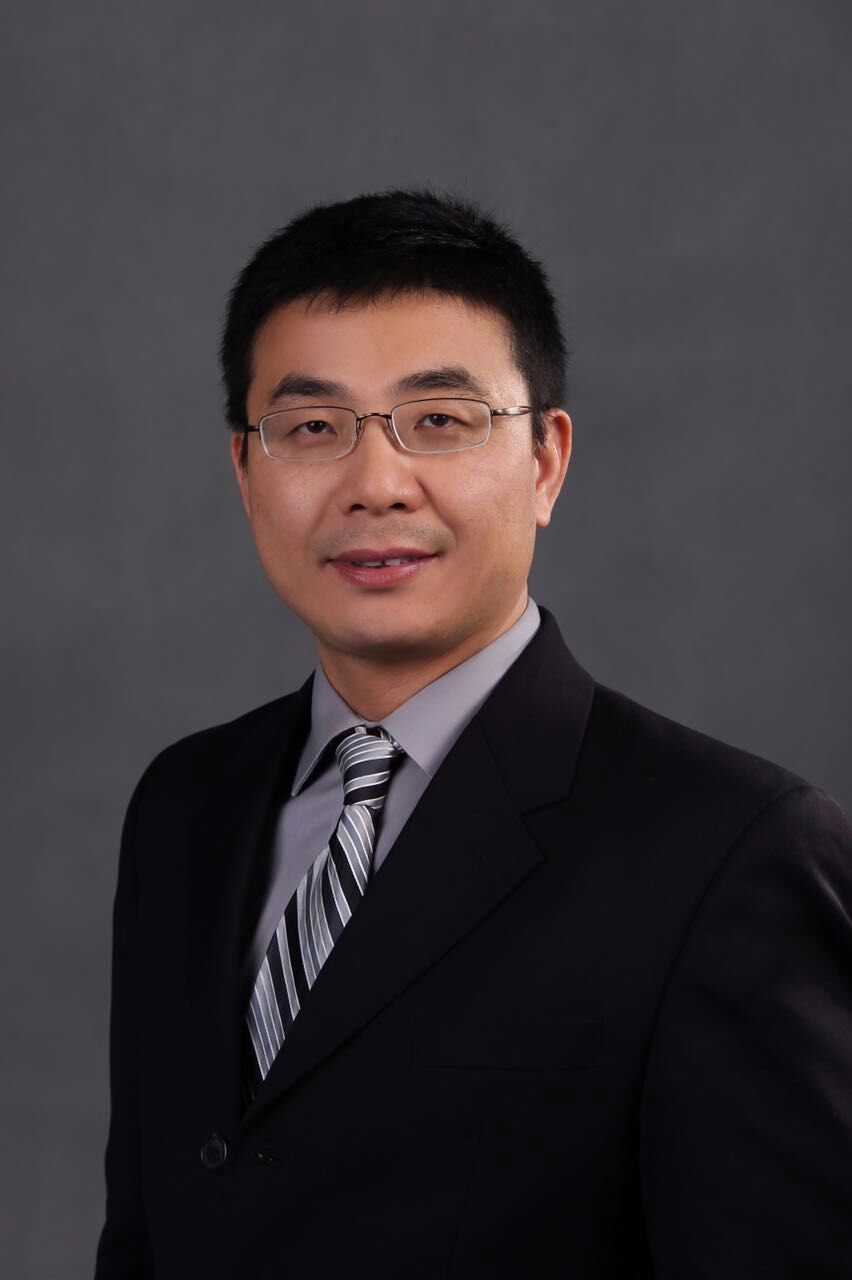}}]
    {Qi Hao} is currently a Professor at the Southern University of Science and Technology, Shenzhen, China. 
    He received the B.E. and M.E. degrees from Shanghai Jiao Tong University, Shanghai, China, in 1994 and 1997, respectively, and the Ph.D. degree from Duke University, Durham, NC, USA, in 2006, all in electrical and computer engineering.
    From 2007 to 2014, he was an Assistant Professor at the Department of Electrical and Computer Engineering, University of Alabama, Tuscaloosa, AL, USA. His current research interests include Intelligent unmanned systems, Machine learning, and Smart sensors.
\end{IEEEbiography}

\vspace{-13mm}
\begin{IEEEbiography}
    [{\includegraphics[width=1in,height=1.25in,trim={25 0 45 0}, clip, keepaspectratio]{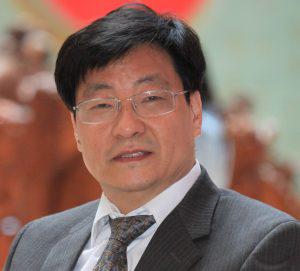}}]
    {Chengzhong Xu (Fellow, IEEE)} is currently a Chair Professor of Computer Science at the University of Macau, Macau, China. He was on the faculty of Wayne State University, Detroit, MI, USA.
    He published two research monographs and more than 500 journal and conference papers with about 20 000 citations, and was a Best Paper Awardee or Nominee of conferences, including HPCA’2013, HPDC’2013, Cluster’2015, ICPP’2015, GPC’2018, UIC’2018, and SoCC’2021. 
    He was also a co-inventor of more than 120 patents and a Co-Founder of Shenzhen Institute of Baidou Applied Technology. 
    He serves or served on many journal editorial boards, including IEEE TC, IEEE TCC, and IEEE TPDS.
    His recent research interests are autonomous driving, cloud and distributed computing, systems support for AI. 
\end{IEEEbiography}

\end{document}